\documentclass[Afour,sageh,times]{sagej}

\usepackage{moreverb,url}

\usepackage[colorlinks,bookmarksopen,bookmarksnumbered,citecolor=red,urlcolor=red]{hyperref}
\usepackage{tasks}
\usepackage{subcaption}
\usepackage{float}
\graphicspath{{Figs/}}
\usepackage{algorithm2e}
\usepackage[noend]{algpseudocode}
\usepackage[colorinlistoftodos]{todonotes}
\usepackage{appendix}
\usepackage{hyperref}

\newcommand\BibTeX{{\rmfamily B\kern-.05em \textsc{i\kern-.025em b}\kern-.08em
T\kern-.1667em\lower.7ex\hbox{E}\kern-.125emX}}

\newcommand\widthy{0.47\linewidth}
\newcommand\widthyt{0.77\linewidth}
\def\volumeyear{2022}

\begin{document}

\runninghead{Cox, Beskok, and Hurmuzlu}

\title{Magnetically Actuated Millimeter-Scale Biped}

\author{Adam Cox\affilnum{1}, Sinan Beskok\affilnum{1}, and Yildirim Hurmuzlu\affilnum{1}}

\affiliation{\affilnum{1}Southern Methodist University, USA}

\corrauth{Yildirim Hurmuzlu, Systems Lab,
Southern Methodist University,
Dallas, Texas, USA.}

\email{hurmuzlu@lyle.smu.edu}

\begin{abstract}
This paper introduces a new approach to studying bipedal locomotion. The approach is based on magnetically actuated miniature robots. Building prototypes of bipedal locomotion machines has been very costly and overly complicated. We demonstrate that a magnetically actuated 0.3~gm robot, we call Big Foot, can be used to test fundamental ideas without necessitating very complex and expensive bipedal machines. We explore analytically and experimentally two age old questions in bipedal locomotion: 1. Can such robots be driven with pure hip actuation. 2. Is it better to use continuous or impulsive actuation schemes.

First, a numerical model has been developed in order to study the dynamics and stability of a magnetically actuated miniature robot. We particularly focus on stability and performance metrics. Then, these results are tested using Big Foot. Pure hip actuation has been successful in generating gait on uphill surfaces. In addition, complex tasks such as following prescribed gait trajectories and navigating through a maze has been successfully performed by the experimental prototype. The nature and timing of hip torques are also studied. Two actuation schemes are used: Heel Strike Actuation and Constant Pulse Wave Actuation. With each scheme, we also vary the time duration of the applied magnetic field. Heel Strike actuation is found to have superior stability, more uniform gait generation, and faster locomotion than the Constant Pulse Wave option. But, Constant Pulse Wave achieves locomotion on steeper slopes.

\end{abstract}


\maketitle

\section{Introduction}

 The earliest studies of bipedal walkers began with inverted pendulum models. As research in bipedal locomotion advanced, investigators faced difficulties due to prohibitively large energy demands and very high component costs when attempting to develop experimental prototypes.  Complex numerical models of bipedal walkers have been created to better understand the dynamics behind such systems. Major concerns of modelling have been the model discontinuities due to feet impact with the walking surface and stability analyses of these highly nonlinear systems. The main idea of the present paper is to answer key questions regarding stability and actuation of a very simple bipedal walker. In addition, we develop a magnetically actuated miniature robot as an experimental platform to produce a low cost, 3D printed prototype. Using this platform, we apply the results of the theoretical analyses to a real system and demonstrate its physical utility. To the best of our knowledge such an approach to verifying theoretical results using a magnetically actuated miniature bipedal prototype has never been taken before.

Robotic bipedal locomotion literature is vast, it is impossible to refer to all work performed on this subject in the confined space of the present article. We refer readers to books and survey articles written on the subject and references within (\cite{HURMUZLU2004, Sugihara, GRIZZLE20141955, Murooka, Mikolajczyk, hobbelen2007limit, brogliato_2018}). The past research has been mainly concerned with the study of topics of stability, nonlinear dynamics, and control of robotic bipedal locomotion. In terms of stability, defining periodic gait as a limit cycle and analyzing its stability using Poincar\'e 
sections, Floquet Multipliers, and Lyapunov Exponents has become the standard approach in the field (\cite{Hurmuzlu_Moskowitz_1986,Hurmuzlu_Moskowitz_1987_2,Hurmuzlu_Basdogan_1994,Hurmuzlu_Basdogan_Stoianovici_1996,Garcia1998,Goswami1997, Ahan, Bruijn, ekizos_santuz_schroll_arampatzis_2018, Saeed, Hamed, Dingwell}). 

Devising an actuation scheme for a bipedal robot is not a  straightforward task. One of the obvious approaches is to place electric/hydraulic motors and sensors at the joints. Yet, this renders the robots excessively heavy and significantly complicates the design. In addition, it results in very high energy requirements. Passive locomotion (\cite{McGeer1990}), inspired by earlier works in ballistic walking (\cite{McMahon&Mochon1980}) has emerged as a promising idea to simplify the design and reduce energy demand. These mechanisms produce gait without joint actuation. They can walk down an inclined plane utilizing gravitational energy. Yet, these bipeds have extremely limited applicability, since they can only walk down inclined planes. In addition, it is very well known that they have extremely limited stability range. Nevertheless, researchers have used the basic concept of passive walking in order to generate active walkers (\cite{Tavakoli_Hurmuzlu_2009,Tavakoli_Hurmuzlu_2010,Tavakoli_Hurmuzlu_2013b,Goswami1997,Garcia1998,Spong2005}). Despite these efforts, the presently known walking robots are extremely complex and have very high price tags (Honda's Asimo (\cite{mauk_2023}), Boston Dynamics's Atlas (\cite{dynamics_2023}), Biped Robotics Lab at Michigan's Cassie Blue (\cite{grizzle_2021}), Engineered Arts' Ameca (\cite{arts_limited_2022}), and NASA's R5 (\cite{hall_2015})). Consequently, it has become very difficult for the majority of the researchers in the field to experiment with physical prototypes.

3-D printed miniature robots can offer an attractive option to develop and experiment with low cost bipeds using rapid prototyping methods. One can experimentally study the dynamics and stability of these systems, and develop control methods and strategies to regulate their motion. Miniature walking robots have been produced before using soft and flexible actuators. \cite{Ijaz2020} developed a simple walking robot that uses embedded magnets to flex the robot's abdomen to produce locomotion. Baisch et al. (\cite{Baisch2014,Baisch2011,Baisch2011v2,Baisch2010}) created a family of walking robots called the Harvard Ambulatory MicroRobot (HAMR). These microrobots use custom machined piezoelectric boards that flex with an applied voltage. HAMR was inspired by cockroaches. Miniaturizing actuators for rigid body robots is difficult. \cite{lynne_1996} provide ideas for miniaturizing actuators. But, for actuators that perform like electric motors, often these systems use compliant surfaces to allow motion (\cite{Buzzin_2022,Fettweis_2021}). The full extent of these "muscle-like" walking robots can be seen in the LCE-microrobots from \cite{Hao2018}. These microrobots are basically a single torsional muscle, whose design has no other control or means of locomotion. With rigid actuation, \cite{Islam2022} developed a minimally actuated, rigid bipedal walker, including five 3D printed rigid bodies and a single actuator per leg. Each actuator is controlled via an open-loop sinusoidal profile, thereby eliminating the need for feedback. Their reasonably small robot (15 cm) was able to walk on level surfaces and produce turns.

Magnetic actuation  of miniature robots has many useful applications in robotic surgery, drug delivery, small scale manufacturing, and other related engineering topics (\cite{Sitti_Microrobot}). Use of electromagnetic fields as a form of robotic actuation allows untethered, low-power mechanisms to regulate the motion of robots. \cite{Yesin2006} used a combination of Helmholtz and Maxwell coils to control cubed microrobots in one-dimension, followed by \cite{Choi2009} in two-dimensions and eventually \cite{Jeong2010} in three-dimensions. \cite{Kim2013} reduced the number of coils needed for two-dimensional control by  using a pair of coils similar to Helmholtz coils that could receive independently controlled current. In addition, \cite{Mahoney2012} explored various magnetic methods of microrobots with helical screw propulsion and rolling locomotion. \cite{Li2020} created a completely rigid, two-legged magnetic microrobot that uses periodic magnetic fields to induce pivot walking to indirectly manipulate cell aggregates, though it is limited to this one mode of locomotion. While not bipedal, \cite{Khatib2020} developed a rectangular millirobot capable of various modes of locomotion, including pivot-walking, tapping, and tumbling, with high maneuverability in tight spaces by capitalizing purely on the torque created by the magnetic field. They also presented "stag beetle" and "carbot" designs for improved applicability. A special novelty of magnetic actuation is that these robots are powered externally, but also untethered. This removes the need for power storage (batteries, compressed air tanks, hydraulic and pneumatic lines, and any other traditional power source.) or electronics (sensors and controllers). This allows for novel miniaturization. All that is required to power the robots on the body itself is a magnet. This opens the door for rapid prototyping as simpler designs allow better analysis means.


In this paper we present the first magnetically actuated bipedal walking robot. To the best of our knowledge, this robot is also the world’s smallest bipedal robot. The robot weighs 0.3g and is 9mm tall. We call this robot the Big Foot.

The actuation of Big Foot is accomplished by simply placing a set of magnets in the upper body of the standard passive walker. Next, a magnetic field applies a moment on the foot-lifting-dynamics to generate gait. The prototype can be produced within hours using 3D printing techniques at a cost of few Dollars.


The manuscript begins with a description of the robot design. Next, the dynamics are presented, including the impact mechanics. Then, the experimental setup is explained. Finally, the numerical and experimental analyses are presented.

\section{Robot Design}

\begin{figure}[]
\centering
\includegraphics[width=\columnwidth]{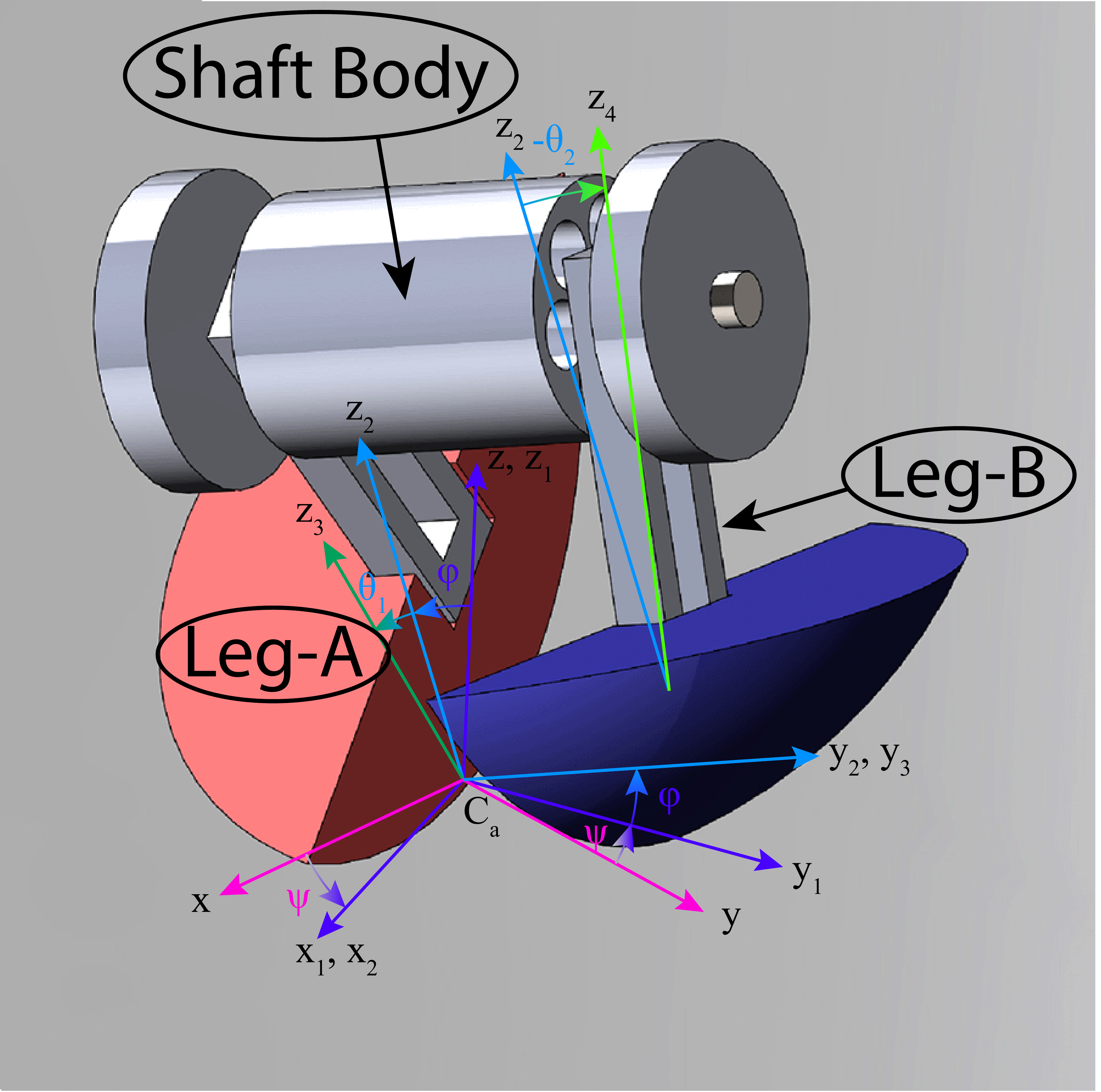}
\caption{Isometric view of the multi-body diagram of Big Foot\label{fig:isometric_MBD}}
\end{figure}  

\begin{figure}[]
\centering
\includegraphics[width=3.5 cm]{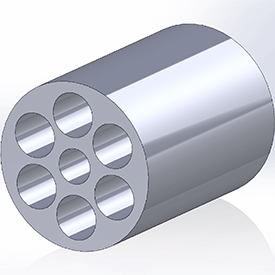}
\caption{Indexed shaft cylinder\label{fig:cylinder2}}
\end{figure}

The central features of Big Foot are the feet, the Shaft Body, and the magnets. Actuation is achieved by placing magnets in the Shaft Body shown in Fig~\ref{fig:isometric_MBD}. A magnetic field is pulsed around Big Foot to create locomotion. The Shaft Body's cylinder is shown in Fig.~\ref{fig:cylinder2}. This cylinder contains outer holes that magnets are press fit into and a central hole that a low friction shaft is pressed into. Coincident to the cylinder are legs. These legs have holes that have a very loose slip fit onto the shaft. There are also collars pressed on the outside of the shaft assembly to prevent disassembly during operation. The cylinder, the legs, and the collars are all made from a very low friction material to allow motion. Finally, at the end of the legs are feet. The purpose of the feet is to enhance stability. Due to the lack of available feedback for such a small walker, stability could only be achieved through physical design. More complicated designs may be considered in future work, but for this version of Big Foot, the curvature of the feet was defined as a circle with a center at the centroid of the Shaft Body. (See Fig.~\ref{fig:Side_2_MBD}).

\section{Robot Dynamics}
\subsection{Coordinate Frames}
A generalized model of Big Foot is shown in Fig~\ref{fig:isometric_MBD} and Fig~\ref{fig:Front_MBD}. The components of Big Foot include two legs with integrated feet, and a central shaft. The legs are designed such that the center of mass is close to the ground, and the feet give Big Foot a vertical static equilibrium orientation.

Six coordinate systems are required to fully model Big Foot's dynamics:
\begin{enumerate}
	\item $\{\boldsymbol{\hat{i}},\boldsymbol{\hat{j}},\boldsymbol{\hat{k}}\}=\boldsymbol{I}$.
	\item $\{\boldsymbol{\hat{i}}_s,\boldsymbol{\hat{j}_s},\boldsymbol{\hat{k}}_s\}=[\boldsymbol{R}_y(\beta)]^T\{\boldsymbol{\hat{i}},\boldsymbol{\hat{j}},\boldsymbol{\hat{k}}\}$.
	\item $\{\boldsymbol{\hat{i}}_1,\boldsymbol{\hat{j}}_1,\boldsymbol{\hat{k}}_1\}=[\boldsymbol{R}_z(\psi)]\{\boldsymbol{\hat{i}}_s,\boldsymbol{\hat{j}}_s,\boldsymbol{\hat{k}}_s\}$.
    \item $\{\boldsymbol{\hat{i}}_2,\boldsymbol{\hat{j}}_2,\boldsymbol{\hat{k}}_2\}=[\boldsymbol{R}_x(\phi)]\{\boldsymbol{\hat{i}}_1,\boldsymbol{\hat{j}}_1,\boldsymbol{\hat{k}}_1\}$.
	\item $\{\boldsymbol{\hat{i}}_3,\boldsymbol{\hat{j}}_3,\boldsymbol{\hat{k}}_3\}=[\boldsymbol{R}_y(\theta_1)]\{\boldsymbol{\hat{i}}_2,\boldsymbol{\hat{j}}_2,\boldsymbol{\hat{k}}_2\}$.
	\item $\{\boldsymbol{\hat{i}}_4,\boldsymbol{\hat{j}}_4,\boldsymbol{\hat{k}}_4\}=[\boldsymbol{R}_y(\theta_2)]\{\boldsymbol{\hat{i}}_2,\boldsymbol{\hat{j}}_2,\boldsymbol{\hat{k}}_2\}$.
\end{enumerate}

The following coordinate transformations are required to fully define the coordinate frames. The simple rotation transformations $ \{[\boldsymbol{R}_x],[\boldsymbol{R}_y],[\boldsymbol{R}_z]\} $ are from Ginsberg's foundational textbook (\cite{ginsberg_2008}). First, a rotation of $\beta$ about $\boldsymbol{\hat{j}}_s$ transforms the Space Fixed coordinate system to the Platform Fixed coordinate system. In Fig~\ref{fig:isometric_MBD} and Fig~\ref{fig:Front_MBD}, the Space Fixed coordinate system is not shown. The Platform Fixed coordinate system is aligned with the surface Big Foot is walking on. Next, a rotation of $\psi$ about $\boldsymbol{\hat{k}}$, followed by a rotation of $\phi$ about $\boldsymbol{\hat{j}}_1$, transforms the Platform Fixed coordinate system to the Shaft Fixed coordinate system. The Platform Fixed coordinate system is chosen as the global for the final equations of motion. Subsequently, a rotation of $\theta_1$ and $\theta_2$ about $\boldsymbol{\hat{j}}_2$ transform the Shaft Fixed coordinate system to the Leg-A and Leg-B coordinate systems. Figure~\ref{fig:isometric_MBD} shows these transformations.

\subsection{Hybrid System}

As a biped robot, Big Foot is a hybrid system (\cite{HURMUZLU2004}). In general for bipedal walkers, there are two states $\{a,b\}$. These states correspond to which foot a bipedal walker is standing on. Big Foot actually has four states corresponding to the contact point between Big Foot and the platform: $\{a,b,c,d\}$ 
\begin{tasks}
\task Leg-A foot inner edge
\task Leg-B foot inner edge
\task Leg-A foot curved surface
\task Leg-B foot curved surface
\end{tasks}

These states are a result of the unique foot design. In numerical simulations, the switching between states is achieved by toggling the respective state to 1 and the other states to 0.

\subsection{Position Vectors}

\begin{figure}[]\centering
	\includegraphics[width=\columnwidth]{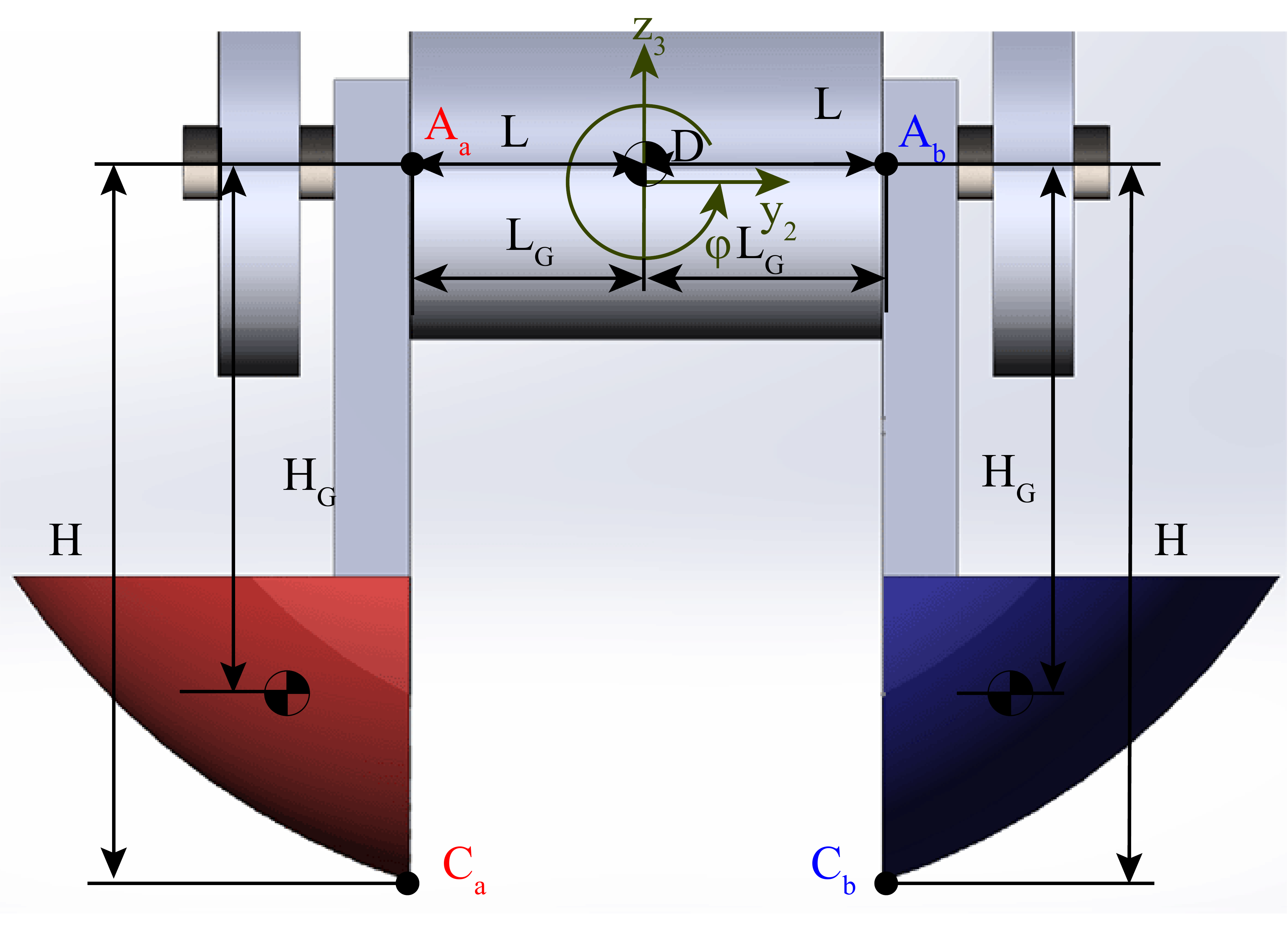}
	\caption{The multi-body diagram of Big Foot from a frontal view\label{fig:Front_MBD}}
\end{figure}

To model the kinematic constraints of motion and the centers of mass, the following position vectors have been defined. These positions are shown in Fig~\ref{fig:Front_MBD}. Point $\boldsymbol{r}_{C_i/O}$ is the position vector of the foot/platform contact point. Point $\boldsymbol{r}_{A_i/O}$ is the position of the center of the joint connecting the leg with the shaft. Finally, $\boldsymbol{r}_{D/O}$ is the position of the center of the shaft.

\begin{equation}\label{key}
	\boldsymbol{r}_{D/O}= \boldsymbol{r}_{A_i/O} + (a-b) L \boldsymbol{\hat{j}}_2.
\end{equation}
where,
\begin{equation}\label{key}
	\boldsymbol{r}_{A_i/O}= \begin{bmatrix}
		X \\
		Y \\
		Z 
	\end{bmatrix}.
\end{equation}

The coordinates of the center of mass for the rigid bodies are as follows. The position vector of the center of mass (COM) of Leg-A is $\boldsymbol{r}_{G_A/O}$ and Leg-B is $\boldsymbol{r}_{G_B/O}$. The COM of the shaft body is $ \boldsymbol{r}_{G_D/O} $. These vectors are defined as:

\begin{align}
\begin{split}
    \boldsymbol{r}_{G_A/O} &=  \boldsymbol{r}_{D/O}-L_G\boldsymbol{\hat{j}}_2-H_G\boldsymbol{\hat{k}}_3. \\ 
	\boldsymbol{r}_{G_B/O} &=  \boldsymbol{r}_{D/O}+L_G\boldsymbol{\hat{j}}_2-H_G\boldsymbol{\hat{k}}_4. \\ 
	\boldsymbol{r}_{G_D/O} &=  \boldsymbol{r}_{D/O}.
\end{split}
\end{align}

\subsection{Angular Velocities}

Next, the angular velocities are defined. The angular velocities of Leg-A, Leg-B, and the Shaft Body are $ \boldsymbol{\omega}_1 $, $ \boldsymbol{\omega}_2 $, and $ \boldsymbol{\omega}_3 $, respectively.

\begin{align}
\begin{split}
    \boldsymbol{\omega}_1 &=  \dot{\psi} \boldsymbol{\hat{k}}+\dot{\phi}\boldsymbol{\hat{i}}_1+\dot{\theta}_1\boldsymbol{\hat{j}}_2. \\ 
	\boldsymbol{\omega}_2 &=  \dot{\psi} \boldsymbol{\hat{k}}+\dot{\phi}\boldsymbol{\hat{i}}+\dot{\theta}_2\boldsymbol{\hat{j}}_2. \\ 
	\boldsymbol{\omega}_3 &=  \dot{\psi} \boldsymbol{\hat{k}}+\dot{\phi}\boldsymbol{\hat{i}}_1.
\end{split}
\end{align}

There is only two angular velocities for the Shaft Body because there are no moments acting about the $ \boldsymbol{\hat{j}}_2 $ direction, which makes one degree of freedom of the Shaft Body static and able to be ignored.

\subsection{Velocity Constraint}

\begin{figure}[]
	\centering
	\includegraphics[width=7 cm]{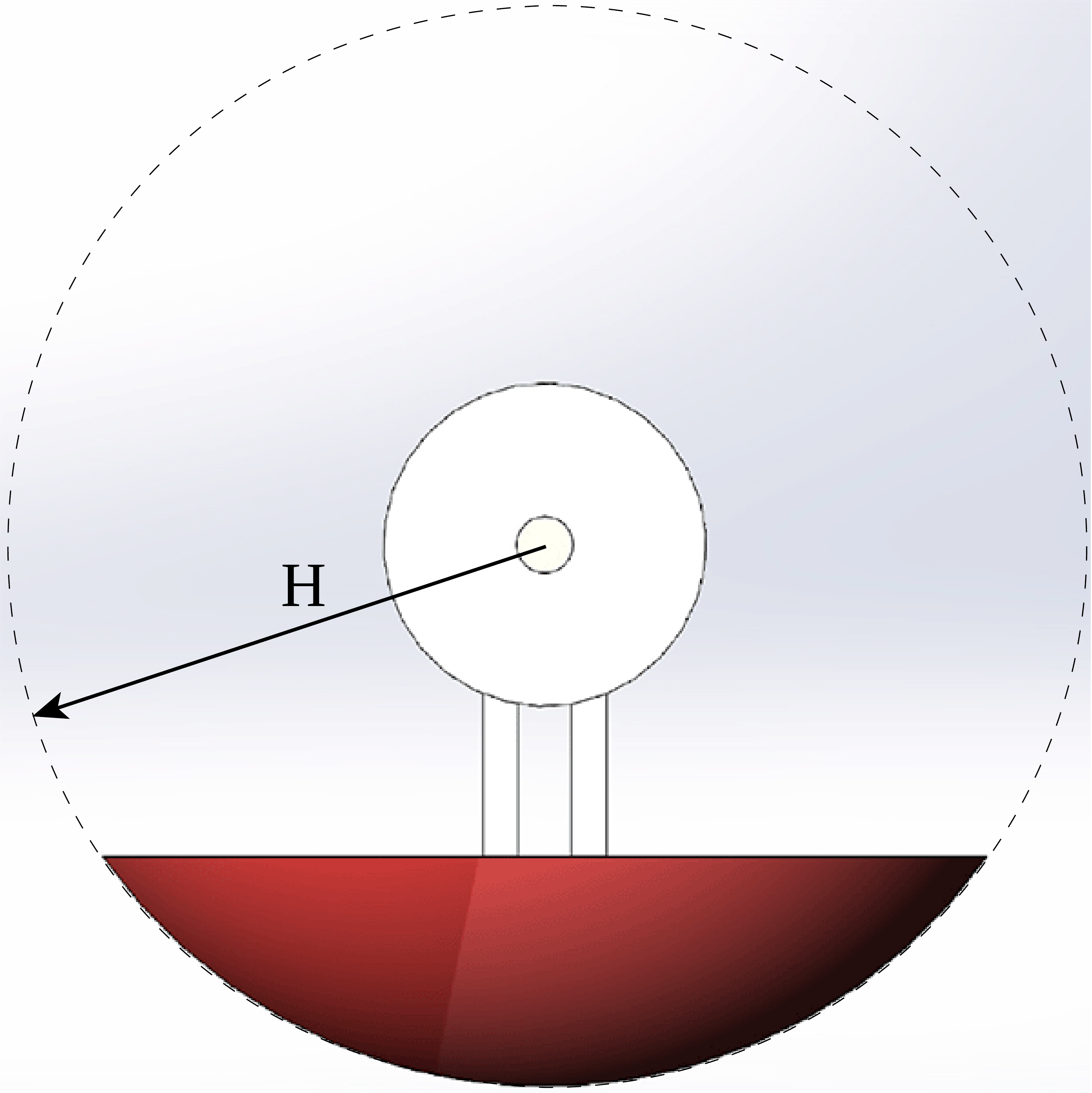}
	\caption{Side view of foot curvature\label{fig:Side_2_MBD}}
\end{figure}

Next, a velocity constraint must be defined for the position of Big Foot with respect to the walking surface. The feet are designed to be circular, as shown in Fig~\ref{fig:Side_2_MBD}.

Two velocity constraints exist for each state. The first velocity constraint is the requirement that the foot is in contact with the ground. This is modeled as the velocity of the foot contact point with the platform, $ \dot{C}_i $, is zero.

Next, a velocity constraint exists that represents the roll of the circular foot on the platform as Big Foot moves. There are four equations for this constraint corresponding to each state.

\begin{tasks}
\task $\boldsymbol{\dot{r}}_{A_i/O}=\boldsymbol{\omega}_1\times\boldsymbol{r}_{A_A/C_A}$
\task $\boldsymbol{\dot{r}}_{A_i/O}=\boldsymbol{\omega}_2\times\boldsymbol{r}_{A_B/C_B}$
\task $\boldsymbol{\dot{r}}_{A_i/O}=\boldsymbol{\omega}_1\times(H_B\boldsymbol{\hat{k}})-L\boldsymbol{\dot{\hat{j}}_2}$
\task $\boldsymbol{\dot{r}}_{A_i/O}=\boldsymbol{\omega}_2\times(H_B\boldsymbol{\hat{k}})+L\boldsymbol{\dot{\hat{j}}_2}$
\end{tasks}
where,
\begin{equation}\label{key}
	\boldsymbol{r}_{A_i/C_i}=H\boldsymbol{\hat{k}}_2.
\end{equation}
and $H_B$ is the radius of the spherical surface of the feet.

The $\boldsymbol{\hat{k}}$-direction of the rolling constraint is holonomic and when integrate for each state is:
\begin{tasks}
\task $Z=H\cos{\phi}$
\task $Z=H\cos{\phi}$
\task $Z=H_B-L\sin{\phi}$
\task $Z=H_B+L\sin{\phi}$
\end{tasks}

\subsection{Energy Definitions}

Now that the position vectors have been fully defined, the Lagrangian can be derived. We start with the different types of energy. First, we have translational kinetic energy:
\begin{equation}
\begin{split}
    T_t=& \frac{1}{2}(m_A\ \boldsymbol{\dot{r}}_{G_A/O}\cdot\boldsymbol{\dot{r}}_{G_A/O}+\\
    & m_B\ \boldsymbol{\dot{r}}_{G_B/O}\cdot\boldsymbol{\dot{r}}_{G_B/O}+m_C\ \boldsymbol{\dot{r}}_{G_C/O}\cdot\boldsymbol{\dot{r}}_{G_C/O})
\end{split}
\end{equation}
where, $m_A$, $m_B$, and $m_C$ are the masses of the two legs and the shaft body, respectively.
Next, the rotational kinetic energy:
\begin{equation}
\begin{split}
    T_r=&\frac{1}{2}(\boldsymbol{\omega}_1^T[\boldsymbol{R}_3]^T \boldsymbol{I}_1 [\boldsymbol{R}_3]\boldsymbol{\omega}_1+\\
    &\boldsymbol{\omega}_2^T[\boldsymbol{R}_4]^T \boldsymbol{I}_2 [\boldsymbol{R}_4]\boldsymbol{\omega}_2+\boldsymbol{\omega}_3^T[\boldsymbol{R}_2]^T \boldsymbol{I}_3 [\boldsymbol{R}_2]\boldsymbol{\omega}_3)
\end{split}
\end{equation}
where, $\boldsymbol{I}_1$ is the inertia matrix of Leg-A with respect to $\{\boldsymbol{\hat{x}}_3,\boldsymbol{\hat{y}}_3,\boldsymbol{\hat{z}}_3\}$, $\boldsymbol{I}_2$ is the inertia matrix of Leg-B with respect to $\{\boldsymbol{\hat{x}}_4,\boldsymbol{\hat{y}}_4,\boldsymbol{\hat{z}}_4\}$, $\boldsymbol{I}_3$ is the inertia matrix of the Shaft Body with respect to $\{\boldsymbol{\hat{x}}_2,\boldsymbol{\hat{y}}_2,\boldsymbol{\hat{z}}_2\}$, and the rotation matrices are defined below:
\begin{align}
\begin{split}
        [\boldsymbol{R}_2]&=\begin{bmatrix}
		\boldsymbol{\hat{i}}_2^T \\
		\boldsymbol{\hat{j}}_2^T \\
		\boldsymbol{\hat{k}}_2^T 
	\end{bmatrix}\\
	[\boldsymbol{R}_3]&=\begin{bmatrix}
		\boldsymbol{\hat{i}}_3^T \\
		\boldsymbol{\hat{j}}_3^T \\
		\boldsymbol{\hat{k}}_3^T 
	\end{bmatrix}\\
	[\boldsymbol{R}_4]&=\begin{bmatrix}
		\boldsymbol{\hat{i}}_4^T \\
		\boldsymbol{\hat{j}}_4^T \\
		\boldsymbol{\hat{k}}_4^T 
	\end{bmatrix}.
\end{split}
\end{align}

Finally, the potential energy of the system is:
\begin{equation}
	V=g(m_A\boldsymbol{r}_{G_A/O}+m_B\boldsymbol{r}_{G_B/O}+m_C\boldsymbol{r}_{G_C/O})\cdot\boldsymbol{\hat{K}}.
\end{equation}

The Lagrangian can be written as:
\begin{equation}
	\mathcal{L}=T_t+T_r-V,
\end{equation}




\subsection{Constraint Forces}

The constrained generalized coordinates for the system are:
\begin{equation}
	\boldsymbol{q}=\{\theta_1, \theta_2, \phi, \psi, x, y\}.
\end{equation}

As a result of the rolling velocity constraint, there are two constraint forces $\{r_x,r_y\}$. These forces interact with the system through generalized forces and moments ($Q_{ic}$):

\begin{align}
\begin{split}
    Q_{1c}=&-(a\ H+c\ H_B \cos{\phi}) (r_x \cos{\psi}+r_y \sin{\psi}) \\
	Q_{2c}=&-(b\ H+d\ H_B \cos{\phi}) (r_x \cos{\psi}+r_y \sin{\psi}) \\
	Q_{3c}=&(r_y\cos{\psi}-r_x\sin{\psi}(H (a + b)\cos{\phi}+\\
	&H_B(c + d) +L(d - c)\sin{\phi}) \\
	Q_{4c}=&(r_x(-\cos{\psi})-r_y \sin{\psi}) (H (a+b) \sin{\phi}+\\
	&L(c-d) \cos{\phi}) \\
	Q_{5c}=&r_x \\
	Q_{6c}=&r_y
\end{split}
\end{align}

There are two types of external loads. The first type of external load is friction. There is rolling friction between the feet and the platform (\cite{Szirtes_2007}):
\begin{align}
\begin{split}
    Q_{1r}&=-c_{r1}sgn(\dot{\theta}_1) \\
	Q_{2r}&=-c_{r1}sgn(\dot{\theta}_2) \\
	Q_{3r}&=-c_{r2}sgn(\dot{\phi})\\
	Q_{4r}&=-c_{r2}sgn(\dot{\psi})
\end{split}
\end{align}
where, $c_{ri}$ is the rolling coefficient for the respective dynamics.

The model also includes the normal force between the foot and the platform, but this normal force varies insignificantly and can be assumed to be constant.

The damping friction term is:
\begin{align}
\begin{split}
    Q_{1f}&=-c_{f1}\dot{\theta}_1 \\
	Q_{2f}&=-c_{f1}\dot{\theta}_2
\end{split}
\end{align}
where, $c_{fi}$ is the damping coefficient for the respective dynamics. The damping friction of $\theta_i$ is the result of the constraint forces between the legs and the Shaft Body.

\subsection{Magnetic Moment}

\begin{figure}[]
	\centering
	\includegraphics[width=6 cm]{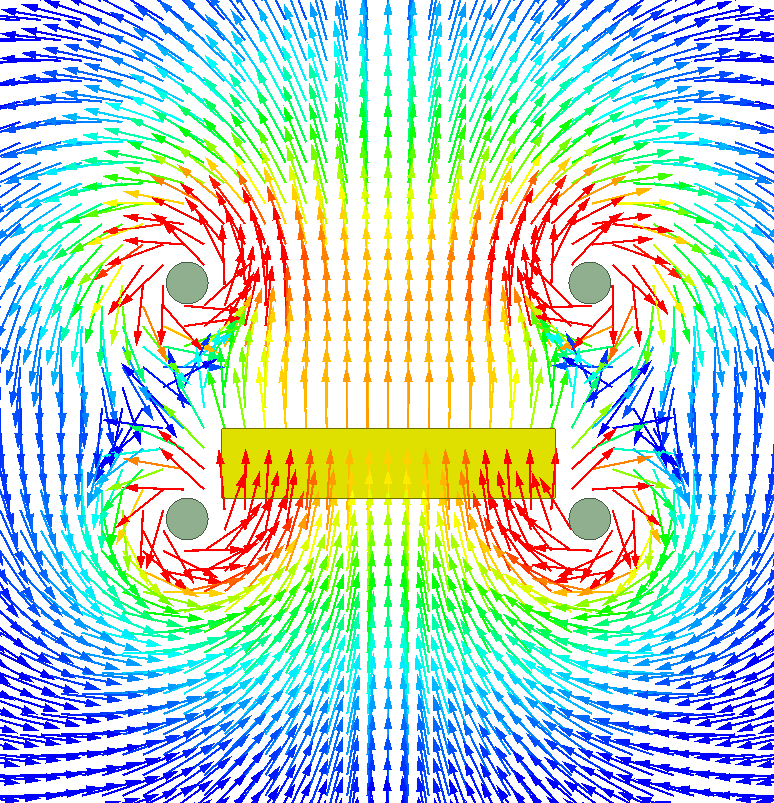}
	\caption{Cross-section view of magnetic field generated by the inner Helmholtz coils.\label{fig:MagneticField2D}}
\end{figure}

The active control of Big Foot is achieved through an arbitrary uniform magnetic field (\cite{Khatib2020}). The magnetic field is generated with 3 sets of orthogonal Helmholtz coils. The magnetic field produced by in the inner set of coils is shown in Fig.~\ref{fig:MagneticField2D}. The yellow rectangle represents the foam platform the experiments are run on. After calibration, the magnetic field can be represented by a vector $p_m \boldsymbol{\hat{j}}_m$, where $p_m$ is the power level of the field and $\boldsymbol{\hat{j}}_m$ is the $y$-direction of the coordinate frame as defined below:
\begin{equation}
    \{\boldsymbol{\hat{i}}_m,\boldsymbol{\hat{j}}_m,\boldsymbol{\hat{k}}_m\}=[\boldsymbol{R}_x(\phi_m)][\boldsymbol{R}_z(\psi_m)]\boldsymbol{\{\hat{i},\hat{j},\hat{k}\}}
\end{equation}

Once the magnetic field has been created, the magnet in Big Foot interacts with the field with the following magnetic moment equation:

\begin{equation}\label{MagneticMoment}
    \boldsymbol{\tau}_m=(k_m\boldsymbol{\hat{j}}_2)\times(p_m\boldsymbol{\hat{j}}_m)
\end{equation}
where, $k$ is a magnet strength coefficient found through tuning the model to experimental results.

Next, to find the applied load of the magnetic moment on the shaft assembly's generalized coordinates, the moment is projected on each coordinate's rotation axis:
\begin{align}
\begin{split}
    Q_{3m}=&\boldsymbol{\tau}_m.\boldsymbol{\hat{i}}_1\\
    =& k_m p_m(\sin{\phi_m} \cos{\phi}-\cos{\phi_m} \sin{\phi} \cos{(\psi_m-\psi)}) \\
	Q_{4m}=&\boldsymbol{\tau}_m.\boldsymbol{\hat{k}}\\
	=& k_m p_m\cos{\phi_m} \cos{\phi} \sin{(\psi_m-\psi)}
\end{split}
\end{align}

Finally, using the Lagrangian method, the equations of motion for Big Foot are:
\begin{align}\label{EquationsOfMotion}
\begin{split}
    \frac{d}{dt}\Bigl(\frac{\partial \mathcal{L}}{\partial \dot{\theta}_1}\Bigr)-\frac{\partial \mathcal{L}}{\partial \theta_1}&=Q_{1c}+(a+c)\ Q_{1r}+Q_{1f} \\
	\frac{d}{dt}\Bigl(\frac{\partial \mathcal{L}}{\partial \dot{\theta}_2}\Bigr)-\frac{\partial \mathcal{L}}{\partial \theta_2}&=Q_{2c}+(b+d)\ Q_{2r}+Q_{2f} \\
	\frac{d}{dt}\Bigl(\frac{\partial \mathcal{L}}{\partial \dot{\phi}}\Bigr)-\frac{\partial \mathcal{L}}{\partial \phi}&=Q_{3c}+Q_{3r}+Q_{3m} \\
	\frac{d}{dt}\Bigl(\frac{\partial \mathcal{L}}{\partial \dot{\psi}}\Bigr)-\frac{\partial \mathcal{L}}{\partial \psi}&=Q_{4c}+Q_{4r}+Q_{4m} \\
	\frac{d}{dt}\Bigl(\frac{\partial \mathcal{L}}{\partial \dot{x}}\Bigr)-\frac{\partial \mathcal{L}}{\partial x}&=Q_{5c} \\
	\frac{d}{dt}\Bigl(\frac{\partial \mathcal{L}}{\partial \dot{y}}\Bigr)-\frac{\partial \mathcal{L}}{\partial y}&=Q_{6c}
\end{split}
\end{align}.

\subsection{Impact Map}
Every time Big Foot switches stance legs, there is an impact. This impact is observed through a loss of kinetic energy and audible sound produced.

\cite{Hurmuzlu2020} notes that the Lagrangian formalism for impact problems can be written as:
\begin{equation}\label{ImpactLagrangian}
\begin{split}
    \Biggl[\frac{\partial T}{\partial \dot{q}_i}\Biggl]^+-\Biggl[\frac{\partial T}{\partial \dot{q}_i}\Biggl]^--\hat{Q}_i-\sum_{K}^{j=1}\lambda_j\frac{\partial\psi_j}{\partial \dot{q}_i}=0\\
    i=1,2,...,N+2
\end{split}
\end{equation}
where, $T$ is the total kinetic energy, the superscripts $-$ and $+$ correspond to the states immediately before and after impact, respectively, $\partial \dot{q}_i$ is the $i$th generalized velocity, $\hat{Q}_i$ is the generalized impulse in the direction of the $i$th generalized coordinate, and $\psi_j$ and $\lambda_j$ are the $j$th constraint and the corresponding Lagrange multiplier, respectively.

In using this impact model, once $\bigl[\frac{\partial T}{\partial \dot{q}_i}\bigr]^+$ and $\bigr[\frac{\partial T}{\partial \dot{q}_i}\bigr]^-$ are evaluated, the velocity constraints are substituted back in. This eliminates the constrained coordinates from the impact model and simplifies the map. Also, to model the movement of the foot/platform contact point of the pre-impact stance leg, the generalized coordinates are expanded as follows:
\begin{equation}
	\boldsymbol{q}_{impacts}=\{\theta_1, \theta_2, \phi, \psi, c_x, c_y, c_z\}.
\end{equation}

The position of these contact points is incorporated by defining the velocity constraint at impact as:
\begin{equation}\label{ImpactVeloConsstraint}
    \boldsymbol{\dot{r}}_{A_i/O}=\boldsymbol{\dot{c}}_i+a(\boldsymbol{\omega}_1\times\boldsymbol{r}_{A_A/C_A})+b(\boldsymbol{\omega}_2\times\boldsymbol{r}_{A_B/C_B})
\end{equation}
where, $\boldsymbol{\dot{c}}_i$ is the velocity of point $c_i$ and state $a$ is used when the stance leg switches from Leg-a to Leg-b and visa versa.

The generalized impulse, $\hat{Q}_i$, is found as:
\begin{align}
    \begin{split}\label{generalizedImpulses}
	Q_{1c}&=(a-b) H (\tau_x\cos{\psi}+\tau_y \sin{\psi})\\
	Q_{2c}&=(b-a) H (\tau_x\cos{\psi}+\tau_y \sin{\psi})\\
	Q_{3c}&=2(a-b)L\ \tau_z \\
	Q_{4c}&=-2(a-b) L (\tau_x\cos{\psi}+\tau_y \sin{\psi})\\
	Q_{5c}&=\tau_x\\
	Q_{6c}&=\tau_y\\
	Q_{7c}&=\tau_z
\end{split}
\end{align}

where, $\tau_i$ is the contact impulses at the striking foot and $\phi$ is set equal to 0 because this is the stance switching angle.

Finally, we have a set of equations to solve. There is the Lagrangian formalization from Eq.~\ref{ImpactLagrangian} with the generalized impulses in Eq.~\ref{generalizedImpulses} and the substituted impact velocity constraints in Eq.~\ref{ImpactVeloConsstraint}. Next, there are post- and pre-impact velocity constraints. Before impact, the contact point on the stance leg is static, meaning:
\begin{equation}
    \boldsymbol{\dot{c}}^-=0
\end{equation}
After impact, the post-impact stance contact point is static. This is represented kinematically as:
\begin{align}
\begin{split}
    \dot{c}_x^+&=-\Bigl((a-b) \cos{\psi} \bigl(H \dot{\theta}_1^+-H \dot{\theta}_2^+-2L\dot{\psi}^+\bigr)\Bigr) \\
	\dot{c}_y^+&=(a-b) \sin{\psi} \bigl(-H\dot{\theta}_1^++H\dot{\theta}_2^++2L\dot{\psi}^+\bigr) \\
	\dot{c}_z^+&=2L(b-a)\dot{\phi}^+
\end{split}
\end{align}
This post impact constraint comes from assuming the coefficient of restitution is 0. Next, these equations are fed into a numerical solver and the post-impact velocities can be found through the solution. This solution is called the impact map.

Once the equations of motion and the impact map has been derived, the system can be simulated.

\section{Actuation Schemes}

We tailor the actuation such that we can study the following questions regarding the application of the magnetic torque:
\begin{enumerate}
    \item The timing of the external torque, whether it should be applied at the end of the heel strike or whether it should be in the form of an external beat and let the biped adjust to it.
    \item The nature of the pulse, whether it should be in the form of high-magnitude short duration (impact like) pulse or longer duration but lower magnitude pulse. 
\end{enumerate}
During  both approaches a magnetic field in the form of square wave is considered. The applied field has five control parameters:
\begin{itemize}
    \item $\psi_m$, is the yaw angle of the magnetic field vector.
    \item $\phi_m$, is the roll angle of the magnetic field vector.
    \item $P_L$, is the time duration of the applied magnetic field.
    \item $P_m$, is the height of the applied magnetic field.
    \item $P_{Area}$ is the magnitude of the applied magnetic impulse
\end{itemize}
where, $P_m$ is unitless. $P_m$ is a multiplier of a magnetic field power defined in the equations of motion.

The rotational impulse acting on the biped by the magnetic field is defined as:
\begin{equation}
    I_m=\int_{t_i}^{t_i+P_L} ||\boldsymbol{\tau}_m|| \,dt
\end{equation}
where, $t_i$ and $t_i+P_L$ are the time of onset and termination of the magnetic pulse respectively. The consequence of considering a square pulse is:
\begin{equation}
        P_{Area}=P_L \times P_m
\end{equation}
with $P_{Area}$, regimes can be defined that correspond to the interaction of the input magnetic field with the dynamical response.
\begin{description}
    \item [I. Impact regime :] This can be described as a Very short duration and a high magnitude pulse. During this regime, the generalized coordinates and velocities remain relatively unchanged during application of the magnetic torque. Thus, the applied magnetic torque has a similar effect to that of an external impact. 
    \item [II. Impulsive regime:] This can be described as a moderate magnitude and duration pulse. During this regime, the generalized coordinates and velocities change significantly during the application of the magnetic torque. Thus, the applied magnetic torque cannot be characterized as an external impact.
\end{description}

\begin{figure}[]
\begin{subfigure}{\linewidth}
        \centering
	\includegraphics[width=\textwidth]{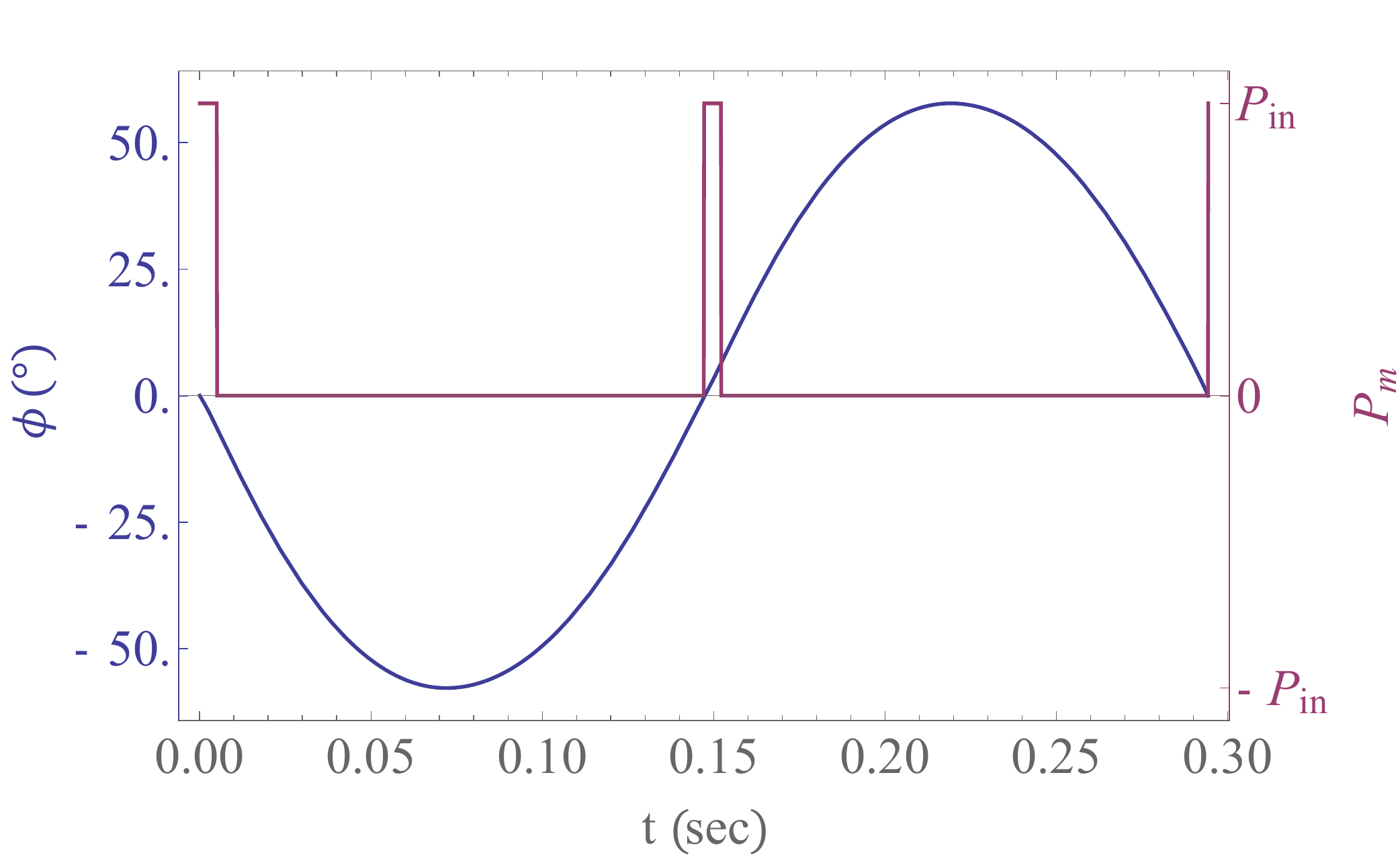}
        \caption{Heel strike actuation scheme in impact regime\label{fig:Short Pulse Length Heel Strike Regime Plot}}
\end{subfigure}
\par\medskip
\begin{subfigure}{\linewidth}
        \centering
	\includegraphics[width=\textwidth]{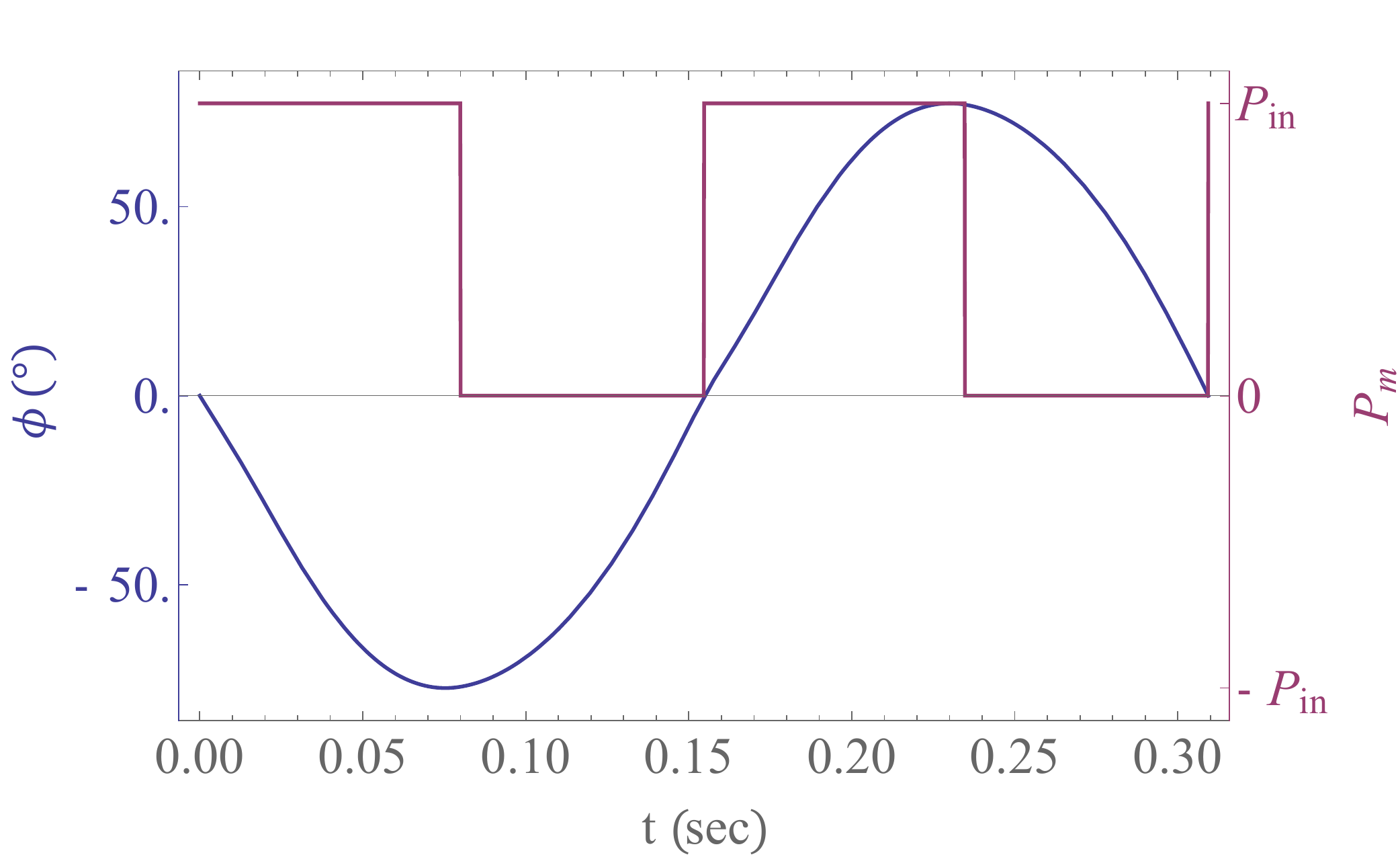}
        \caption{Heel strike actuation scheme in impulsive regime\label{fig:Long Pulse Length Heel Strike Regime Plot}}
\end{subfigure}
\par\medskip
\begin{subfigure}{\linewidth}
        \centering
	\includegraphics[width=\textwidth]{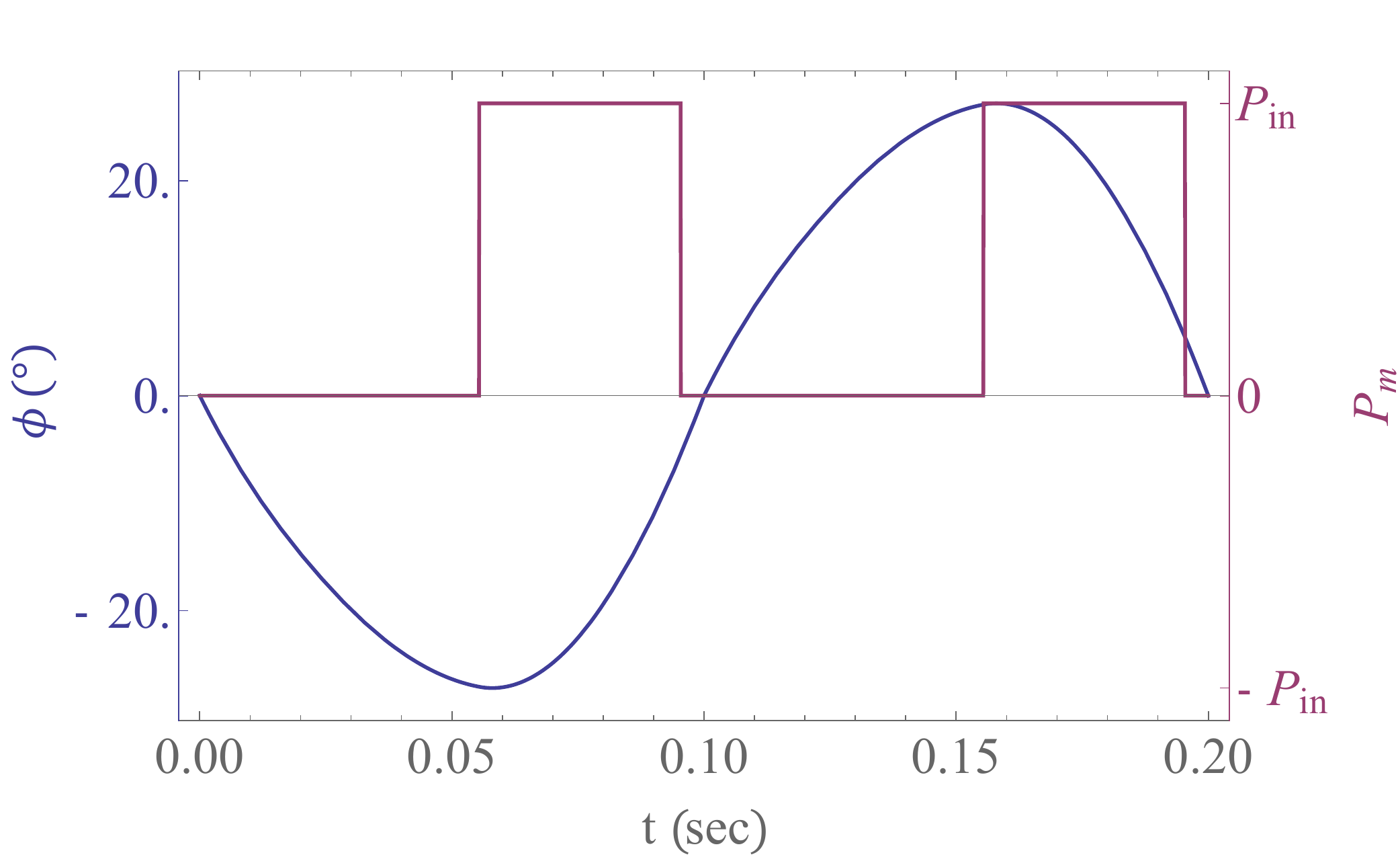}
        \caption{Constant pulse wave actuation scheme in impulsive regime\label{fig:Constant Pulse Wave Regime Plot}}
\end{subfigure}
\caption{The two actuation schemes in different regimes\label{fig:Actuation Schemes}}
\end{figure}

Responses illustrating these regimes are shown in Figs.~\ref{fig:Short Pulse Length Heel Strike Regime Plot} and \ref{fig:Long Pulse Length Heel Strike Regime Plot}.

In addition, as mentioned before, two actuation timing approaches will be considered. 
\begin{description}
    \item[I. Heel strike based actuation:] The first scheme is based on triggering the pulse immediately after the heel strike:

\begin{equation}
    \phi[t]=\begin{cases} 
      \phi_{in} & \phi> 0 \\
      -\phi_{in} & \phi< 0
   \end{cases}\;\;\;\text{,}
\end{equation}

\begin{equation}
    \psi[t]=\begin{cases} 
      \psi_{in} & \phi> 0 \\
      -\psi_{in} & \phi< 0
   \end{cases}\;\;\;\text{, and}
\end{equation}

\begin{equation}
    P_m=\begin{cases} 
      P_{in} & t_s<t<t_s+P_L \\
      0 
   \end{cases}
\end{equation}
where, $t_s$ is the time of heel strike. An example run of this foot strike actuation is shown in Fig.~\ref{fig:Short Pulse Length Heel Strike Regime Plot} and Fig.~\ref{fig:Long Pulse Length Heel Strike Regime Plot}.
\item[II. Constant period actuation]The second actuation scheme is completely open-loop and is simply a pattern of periodically applied pulses:

\begin{equation}
    \phi_m=\begin{cases} 
       \phi_{in} & 0<\text{mod}{(t,t_4)}<t_1 \\
       -\phi_{in} & t_2<\text{mod}{(t,t_4)}<t_3
   \end{cases}\;\;\;\text{,}
\end{equation}

\begin{equation}
    \psi_m=\begin{cases} 
       \psi_{in} & 0<\text{mod}{(t,t_4)}<t_1 \\
       -\psi_{in} & t_2<\text{mod}{(t,t_4)}<t_3
   \end{cases}\;\;\;\text{, and}
\end{equation}

\begin{equation}
    P_m=\begin{cases} 
       P_{in} & (0<\text{mod}{(t,t_4)}<t_1)||(t_2<\text{mod}{(t,t_4)}<t_3) \\
       0
   \end{cases}
\end{equation}
where, $t_1=P_L$, $t_2=P_L+t_{off}$, $t_3=2 P_L+t_{off}$, $t_4=2 P_L+2 t_{off}$, and $t_{off}$ is the time period that the pulse is turned off.

An example run of this constant pulse wave actuation is shown in Fig.~\ref{fig:Constant Pulse Wave Regime Plot}.
\end{description}

In Fig.~\ref{fig:Actuation Schemes}, the plots show steady state responses. Notice the timing between the heel strike and the switching of the input pulse. For the heel strike actuation, as enforced by the magnetic torque input, the heel strike and the input pulse are synchronous. But, for the constant pulse wave, the pulse does not necessarily initiate at the heel strike. This shows how the gait pattern can be initiated without the syncing of the pulse wave and the heel strike. The selection of actuation parameters depend on the desired gait characteristics. 

\section{Stability Analysis}
Let $\phi_p(t)=(\theta_1, \theta_2, \phi , \psi, \dot{\theta}_1, \dot{\theta}_2, \dot{\phi}, \dot{\psi})\in\mathbb{R}^{8}$ be a periodic solution of Eq.~\ref{EquationsOfMotion} (\cite{Kashki_2017}). The one-sided hyper plane Poincar\'{e} section is defined as
\begin{equation}
    \mathcal{S}\equiv(\theta_1, \theta_2, \psi, \dot{\theta}_1, \dot{\theta}_2, \dot{\phi}, \dot{\psi})\in\mathbb{R}^{7}:\phi=0,\dot{\phi}<0.
\end{equation}

If $\boldsymbol{\xi}[k]\in \mathcal{S}$ denotes the $k$th intersection of $\mathcal{S}$ by the flow of $\phi_p$, the discrete-time Poincar\'{e} Map $\mathcal{P}:\mathcal{S}\rightarrow\mathcal{S}$ can be expressed as 
\begin{equation}
    \boldsymbol{\xi}[k+1]=\mathcal{P}(\boldsymbol{\xi}[k])
\end{equation}

Subsequently, if $\boldsymbol{\xi}^*$ stands for a fixed point of the Poincar\'{e} Map, then the local exponential stability of $\boldsymbol{\xi}^*$ on $\mathcal{S}$ is equivalent to local exponential stability of the underlying limit cycle (\cite{westervelt_grizzle_chevallereau_morris_2007}). We used Floquet Theory for stability analysis of a specific limit cycle (\cite{Hurmuzlu_Moskowitz_1987,Hurmuzlu_Moskowitz_1987_2}). Therefore, the local linearization of the Poincar\'{e} Map about $\boldsymbol{\xi}^*$ gives:

\begin{align}
    \begin{split}
        \mathcal{P}(\boldsymbol{\xi})\simeq&\mathcal{J}(\boldsymbol{\xi}^*)(\boldsymbol{\xi}-\boldsymbol{\xi}^*)\\
        \mathcal{J}(\boldsymbol{\xi})=&\frac{\partial\mathcal{P}(\boldsymbol{\xi})}{\partial\boldsymbol{\xi}}
   \end{split}
\end{align}
where $\mathcal{J}(\boldsymbol{\xi})$ is the $7\times7$ linearized Jacobian matrix of $\mathcal{P}(\boldsymbol{\xi})$. Next, the Floquet multipliers are defined as:
\begin{equation}
    \rho_i=|Re(\lambda_i)|:\lambda_i=\text{eig}(\mathcal{J}(\boldsymbol{\xi}))
\end{equation}
where $\rho_i$ and $\lambda_i$ are the $i$th Floquet multiplier and eigenvalue of $\mathcal{J}(\boldsymbol{\xi})$. Accordingly, stability of the limit cycle can be defined as:
\begin{equation}
        \phi_p(t)=\begin{cases} 
      \text{stable}: & \forall\rho_i<1 \\
      \text{unstable}: & \exists\rho_i\ge1
   \end{cases}
\end{equation}

The algorithm used to find the Floquet multipliers is shown below:

\begin{algorithm}
	\caption{Floquet Multiplier Finder}\label{FloquetFinder}
		Define list $\boldsymbol{A}=\{\}$\\
	\For{$n\in \mathcal{S}$}{
		$\mathcal{S}(0)_n=(1+\delta)\mathcal{S}(0)_n$\\
		Integrate for one cycle\\
		Join $\boldsymbol{A}$ and $\frac{\mathcal{S}^*-\mathcal{S}(t_f)}{\delta\mathcal{S}(0)_n}$ to form a new $\boldsymbol{A}$, where $t_f$ is the final time of the cycle.\\
	}
	$\rho_i=||\text{eig}(\boldsymbol{A})_i||$\\
\end{algorithm}

Next, we explore the effect of the input parameter values on the resulting gait patterns.

\section{Simulations}

The simulations were conducted using Wolfram Research Mathematica software. In the next subsections we present the simulation results. We study the dynamic behavior for the two actuation schemes considering the following resulting outcomes:
\begin{enumerate}
    \item Existence and type of resulting limit cycles.
    \item Stability of the period-1 limit cycles.
    \item Step length and progression speed of the biped.
\end{enumerate}
We vary the following parameters when we conduct the simulations:
\begin{enumerate}
    \item Pulse duration $P_L$
    \item Pulse area $P_{Area}$
    \item Walking surface inclination $\beta$
\end{enumerate}
We have two objectives here: a) first, is to identify the parameter ranges that result in the most stable period-1 cycles with the highest possible progression speed over the steepest walking surface, b) second, is to compare the dynamic response for impact and impulsive regimes.

\subsection{Heel Strike Scheme Simulation Results}
\begin{figure}[]
	\centering
	\includegraphics[width=\columnwidth]{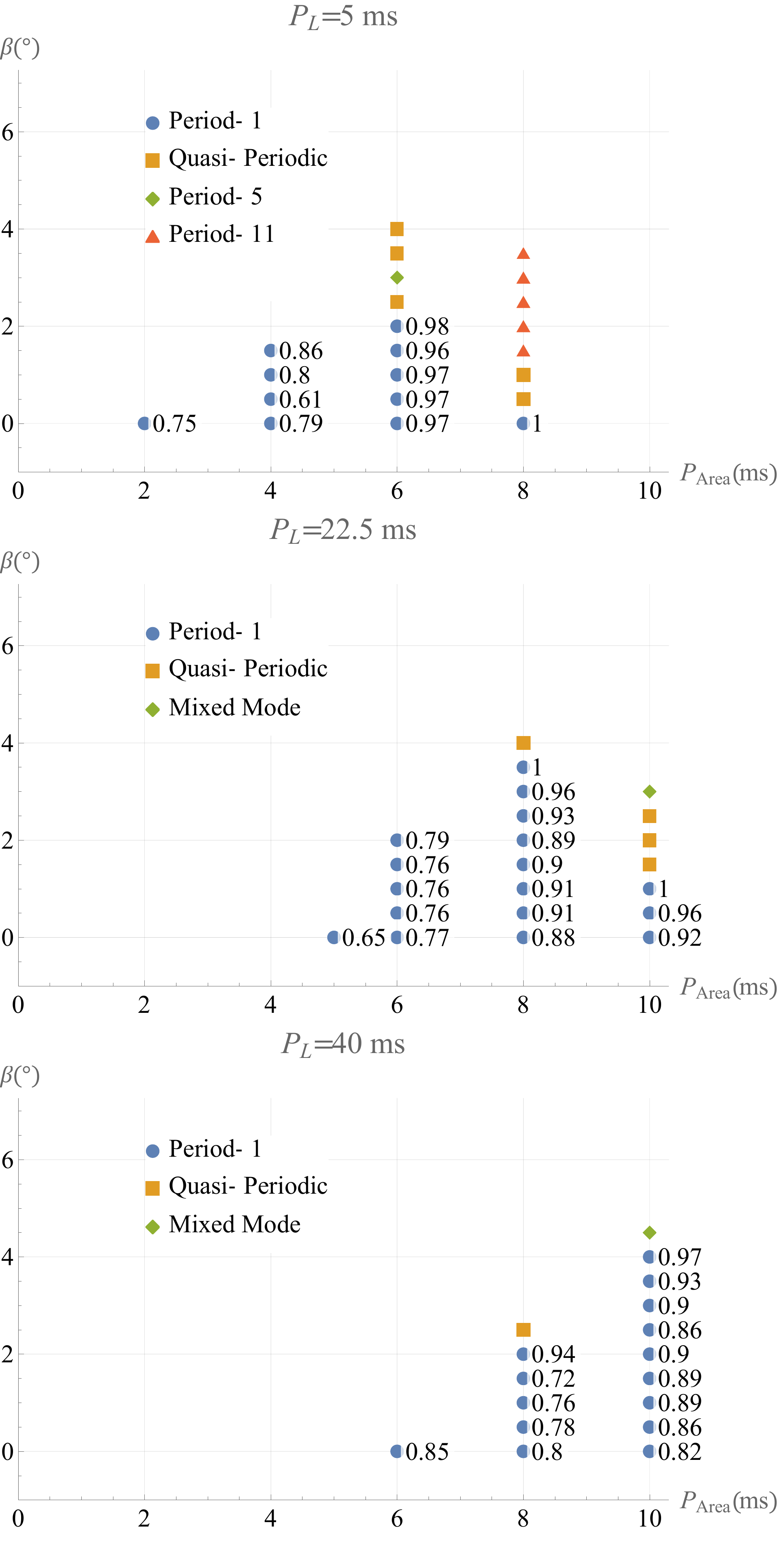}
	\caption{Forward gait patterns in heel strike actuation\label{fig:HeelStrikeBifurcationsWithConstantAreas}}
\end{figure}

In Fig.~\ref{fig:HeelStrikeBifurcationsWithConstantAreas}, the results of the simulations for the heel strike actuation scheme are shown. The results are composed of 3 sets of different $P_L$ values:$\{5\text{ ms}, 22.5\text{ ms}, 40\text{ ms}\}$. In each set five pulse areas were used:$\{2\text{ ms}, 4\text{ ms}, 6\text{ ms}, 8\text{ ms}, 10\text{ ms}\}$. For simplicity, the direction of the applied magnetic pulse is kept constant at $\{\phi_m=60^\circ,\psi_m=20^\circ\}$. The simulations were initiated by running the model at a $0^\circ$ slope for each $P_{Area}$ until either a limit cycle is identified, a pre-determined simulation time has expired, or locomotion is lost. Once the steady state locomotion is identified for a flat slope, the nature of the gait is added to the bifurcation map, the average maximum Floquet multiplier is calculated, and the slope is incremented by $0.5^\circ$ for the next run. The increasing of slope continued until either the robot begins walking backward or locomotion is lost.

In algorithm~\ref{FloquetFinder}, the calculation of the Floquet multipliers is explained. The perturbation, $\delta$, is varied between 0.01 and 0.1. Then, for each $\delta$, the maximum Floquet multiplier is selected. Finally, the average of the maximum Floquet multipliers for each $\delta$ is calculated to represent the average local stability of the fixed point. This value is printed next to the corresponding sample point on Fig.~\ref{fig:HeelStrikeBifurcationsWithConstantAreas} and Fig.~\ref{fig:ConstantPulseWaveBifurcations_60}. From the map, the maximum slope the robot can walk and the corresponding gait patterns and stability has been identified.

As stated  previously, we conduct the simulations in order to identify the optimal operating region for Big Foot. We have chosen three goals for locomotion. The first goal is to walk the steepest slope. Based on our results, longer pulse lengths with higher pulse areas is best (larger $P_m$). In our runs, a maximum slope of $4.5^\circ$ at a pulse length of $40\text{ ms}$ and a pulse area of $10\text{ ms}$ is achieved. The second goal is to walk the steepest slope with a period-1 gait. This occurred at $4^\circ$ at the same parameter values. The period-1 gait is useful for controllability as the displacement per step remains unchanged during locomotion. The third goal is to achieve maximum stability. Based on the average maximum Floquet multipliers, smaller pulse areas result in higher stability.

\begin{figure}[]
	\centering
	\includegraphics[width=0.4\textwidth]{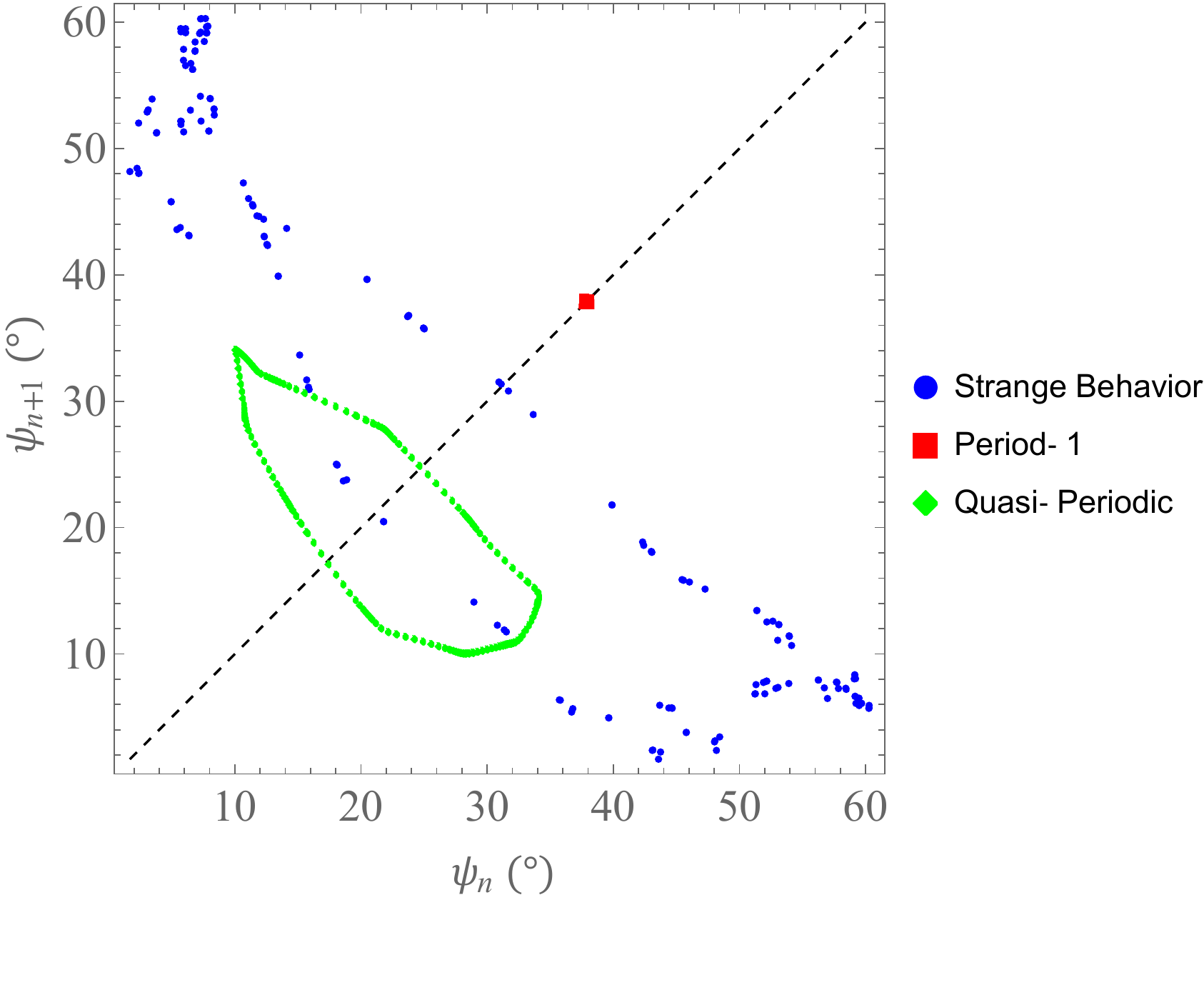}
	\caption{Poincar\'{e} Return Map for the yaw angle in a Quasi-Periodic gait.\label{fig:Quasi-Periodic Return Map}}
\end{figure}

Several overall trends in the results were observed as well. Quasi-Periodic gaits were observed as, while increasing the slope, the gait transitioned from one periodic gait to another. In Fig.~\ref{fig:Quasi-Periodic Return Map}, the Poincar\'{e} return map for $\psi$ on a Quasi-Periodic orbit is shown as an example. For $P_L=5\text{ ms}$ and $P_L=22.5\text{ ms}$, the bifurcation plot have similar trends as $P_{Area}$ is increased. Initial, for $P_L=5\text{ ms}$ and $P_L=22.5\text{ ms}$, as $P_{Area}$ is increased, the maximum slope increases. The maximum for these ranges of $P_{Area}$ is observed as the point to which Big Foot would start walking backwards. For $P_L=5\text{ ms}$ at $P_{Area}=6\text{ ms}$ and for $P_L=22.5\text{ ms}$ at $P_{Area}=8\text{ ms}$, this point switches to being the point of instability. Meaning, for $P_{Area}$ greater than this point, the maximum slope is limited by stability. At these $P_{Area}$, Big Foot falls and loses locomotion at steeper slopes.

\subsection{Constant Pulse Wave Scheme Simulation Results}
\begin{figure}[]
	\centering
	\includegraphics[width=\columnwidth]{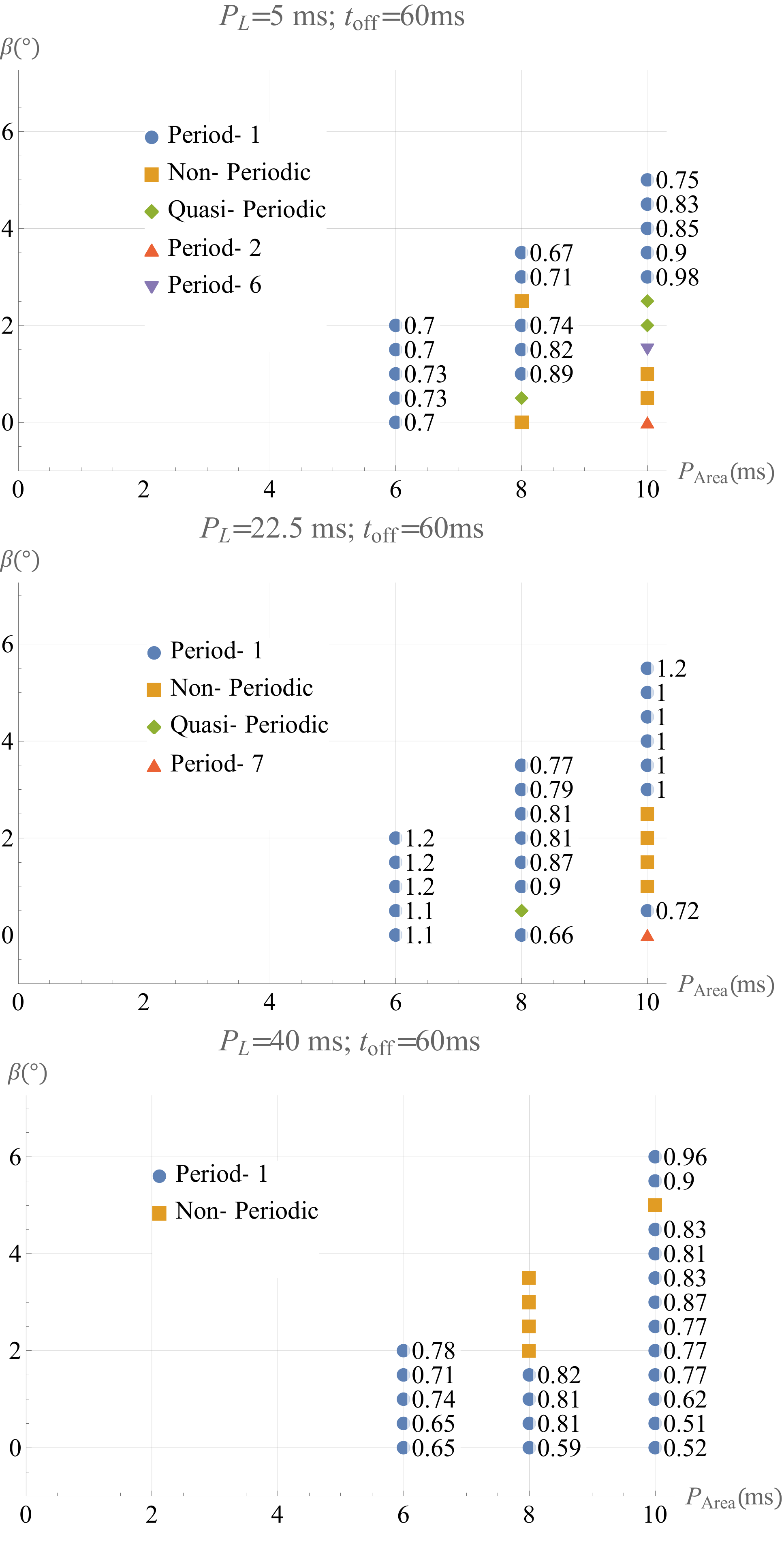}
	\caption{Forward gait patterns in constant pulse wave actuation\label{fig:ConstantPulseWaveBifurcations_60}}
\end{figure}

Next, simulations for the constant pulse wave actuation scheme were conducted. A bifurcation map of the same structure of Fig.~\ref{fig:HeelStrikeBifurcationsWithConstantAreas} is setup for the constant pulse wave scheme. The results of this map are shown in Fig.~\ref{fig:ConstantPulseWaveBifurcations_60}. For this map, the same parameters of Fig.~\ref{fig:HeelStrikeBifurcationsWithConstantAreas} with a $t_{off}=60\text{ ms}$ is chosen.

The difference in results of the heel strike scheme and the constant pulse wave scheme are as follows. The maximum slope achieved by the constant pulse wave scheme is significantly higher with $6^\circ$ at $P_L=40\text{ ms}$ and $P_{Area}=10\text{ ms}$. Another difference is the generation of different periodic gaits. The heel strike scheme predictably performed period-1 gaits at lower slopes. Whereas, the gait patterns of the constant pulse wave do not appear to be predictable from this study. For the constant pulse wave scheme a gait pattern called "Non-Periodic" is identified. "Non-Periodic" refers to a gait that appeared to have a random gait pattern and could not be classified as periodic or quasi-periodic. Another observation is that the performance did not vary greatly with different pulse lengths for the constant pulse wave, where it does for the heel strike. For the constant pulse wave, the robot is only operational in pulse areas $6\text{ ms}$ through $10\text{ ms}$ for all pulse lengths. For a control engineer, the heel strike appeared to be more desirable period-1 gait, but the constant pulse wave scheme achieves significantly greater slopes. Another difference is stability. For the constant pulse wave with $P_L=22.5\text{ ms}$, lower pulse areas were found to be less stable. This is believed to be a result of the same dynamics that cause the lack of predictability in gait pattern as the generation of different gait patterns developed at instability.

\begin{figure}[]
	\centering
	\includegraphics[width=\columnwidth]{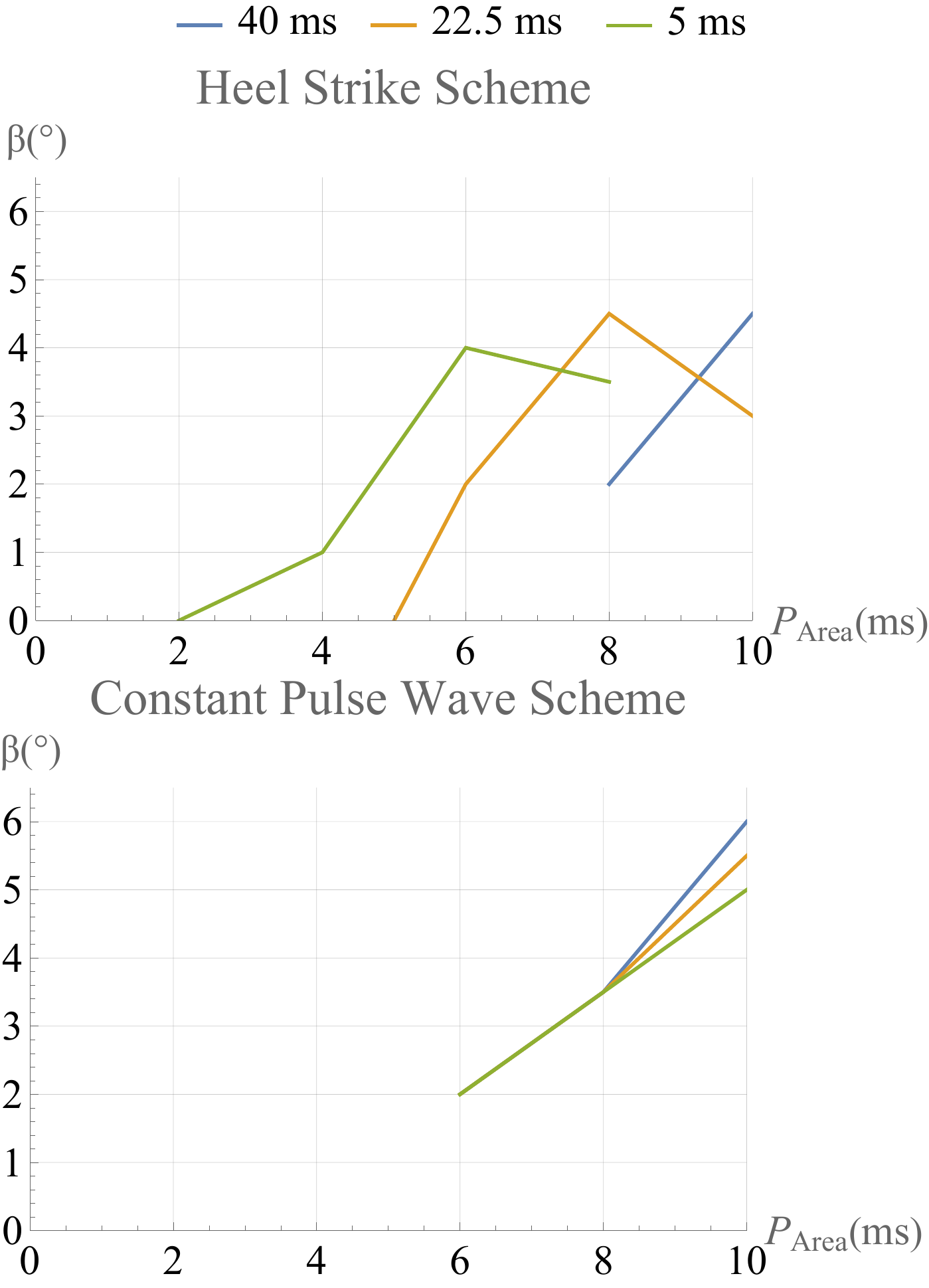}
	\caption{Maximum Slopes Achieved\label{fig:MaxSlopesComposite}}
\end{figure}

In Fig.~\ref{fig:MaxSlopesComposite}, the maximum slopes achieved in the simulations are shown. These plots show the behavior of the performance as you change actuation schemes and input parameters. For the heel strike scheme, the maximum achieved slope appears to increase as you increase the pulse length. However, for the constant pulse wave actuation, the maximum slope occured at $P_{Area}=10\text{ ms}$ for all pulse durations. The maximum achieved slope for the constant pulse wave scheme increased with larger pulse areas.

\begin{figure*}[h!]
\begin{subfigure}{\widthy}
        \centering
	\includegraphics[width=\widthyt]{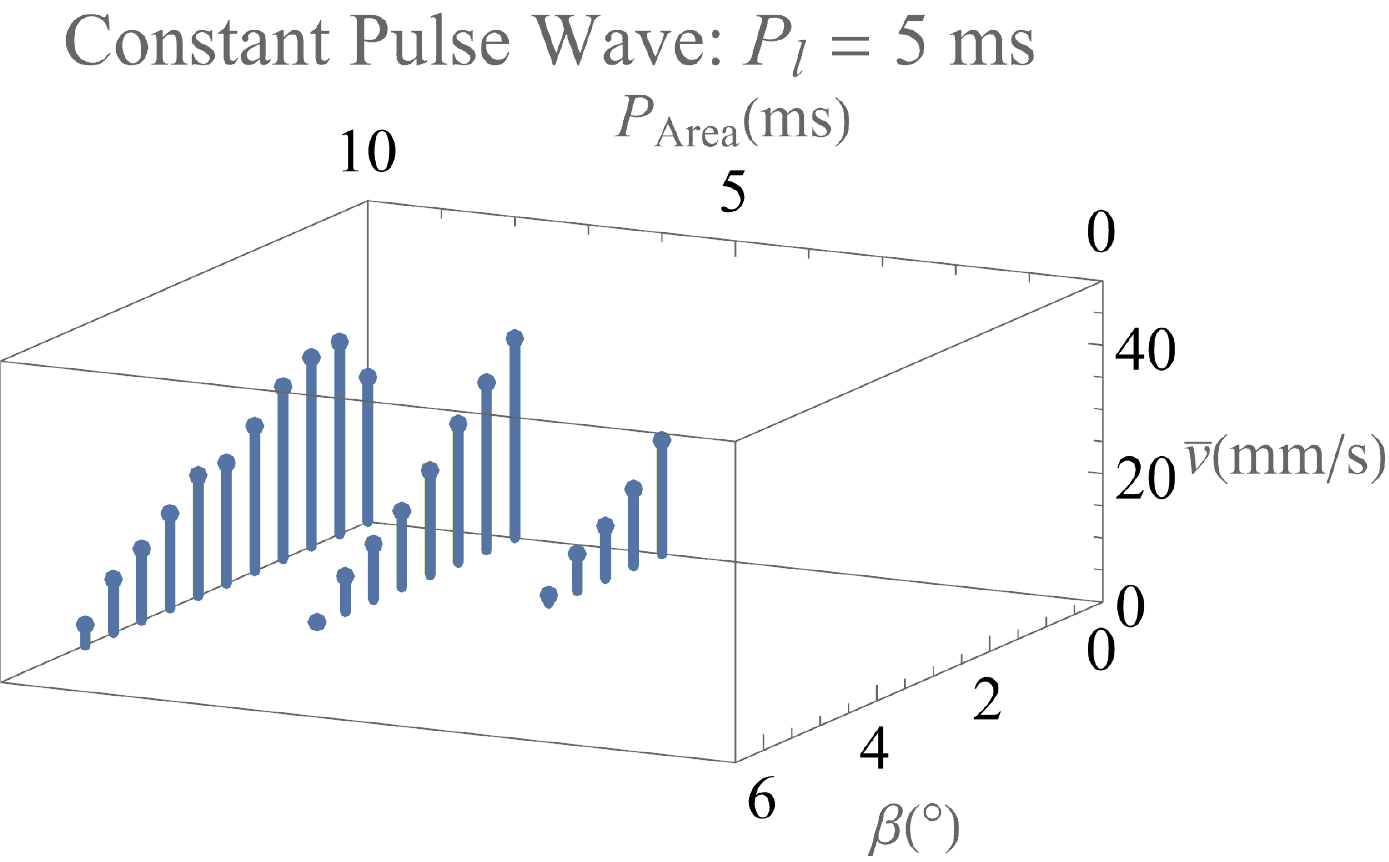}
\end{subfigure}
\begin{subfigure}{\widthy}
        \centering
	\includegraphics[width=\widthyt]{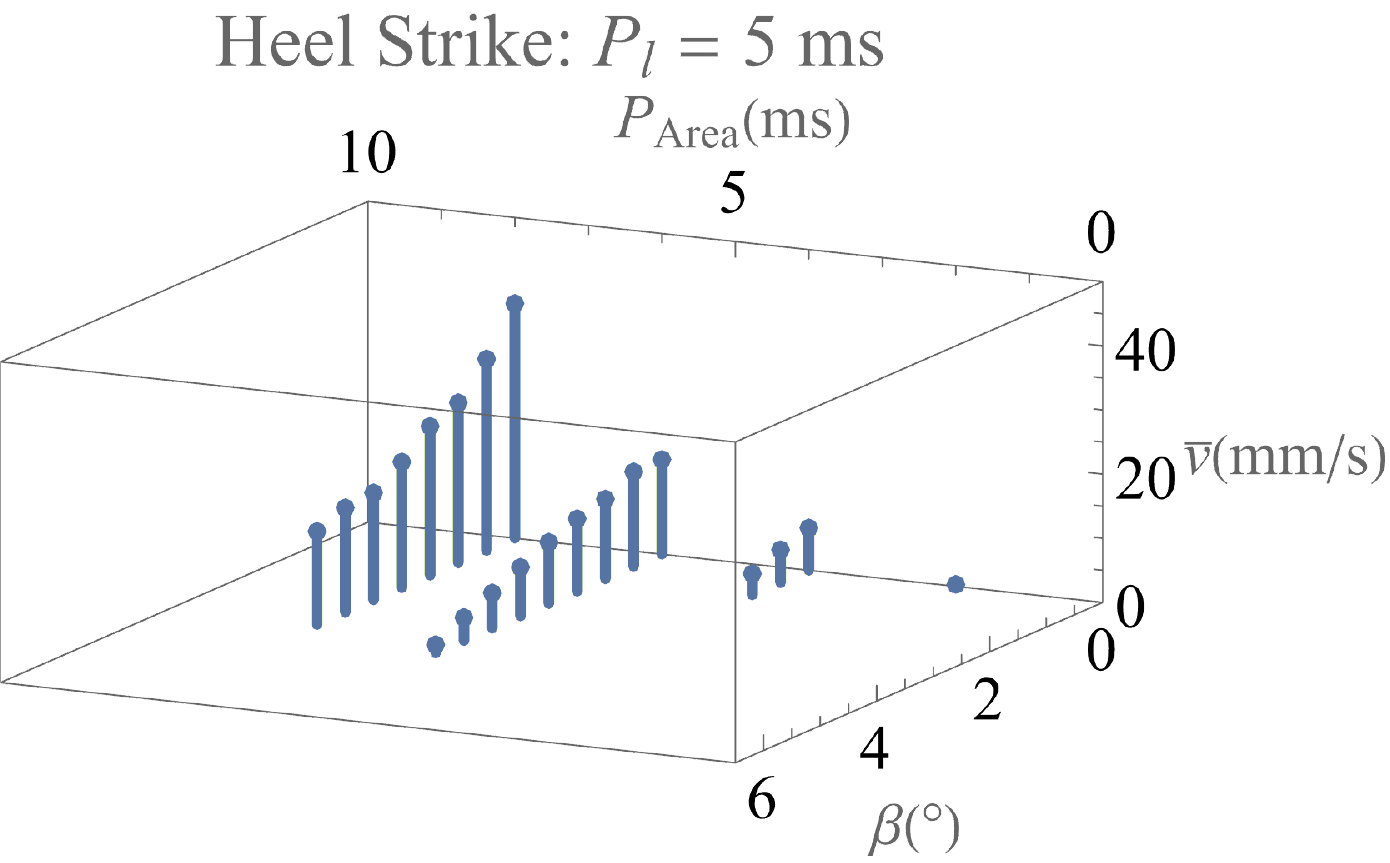}
\end{subfigure}
\begin{subfigure}{\widthy}
        \centering
	\includegraphics[width=\widthyt]{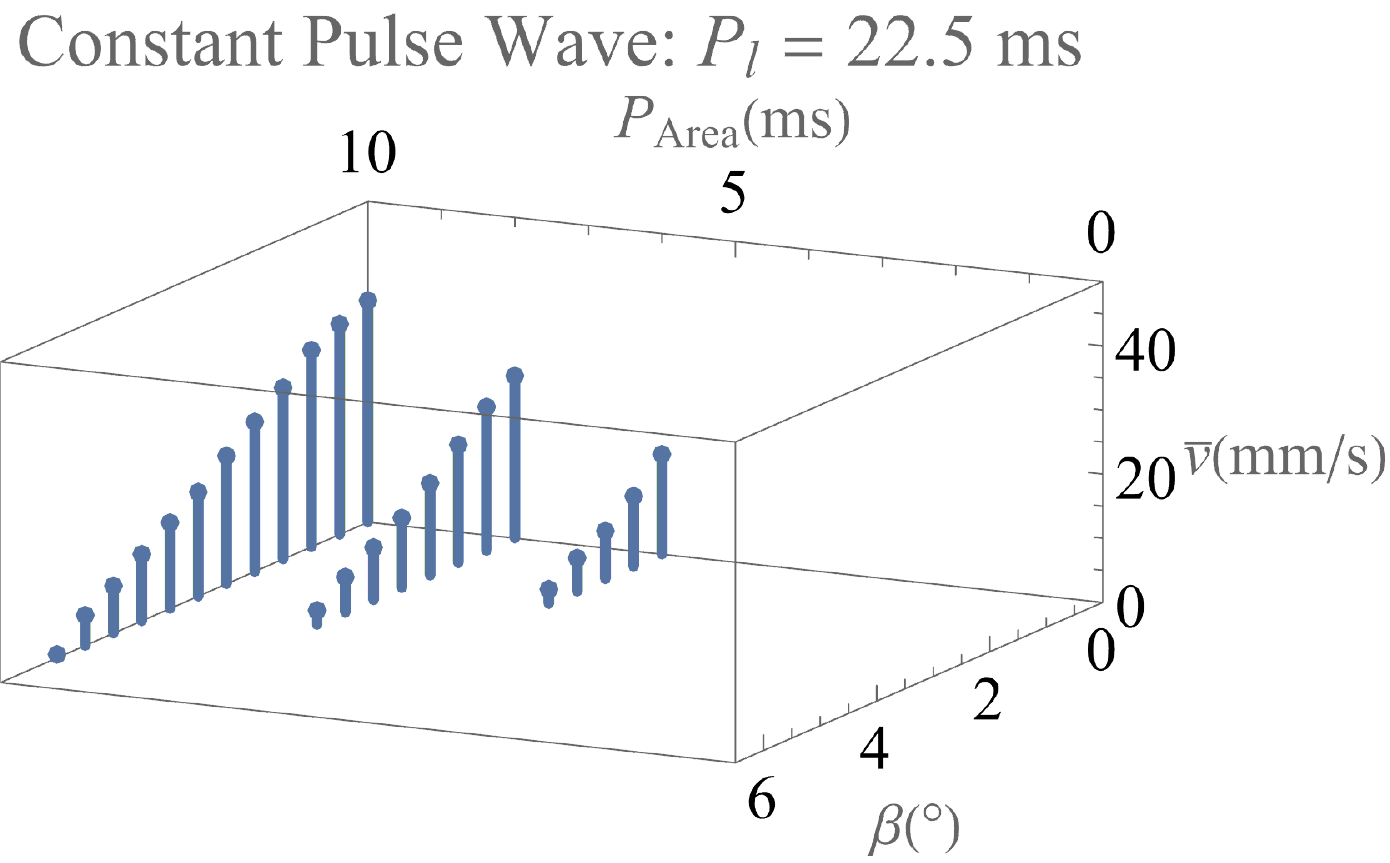}
\end{subfigure}
\begin{subfigure}{\widthy}
        \centering
	\includegraphics[width=\widthyt]{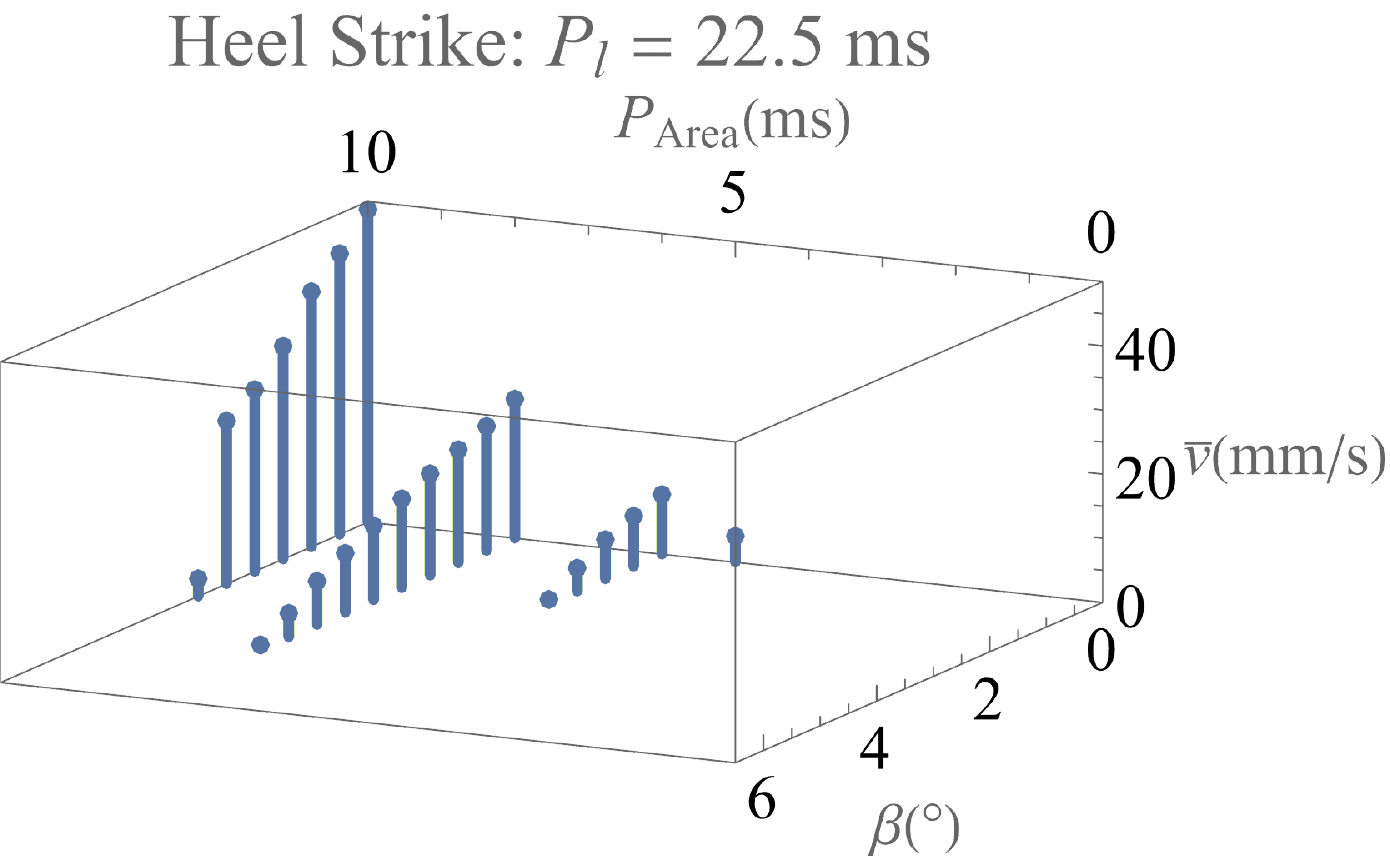}
\end{subfigure}
\begin{subfigure}{\widthy}
        \centering
	\includegraphics[width=\widthyt]{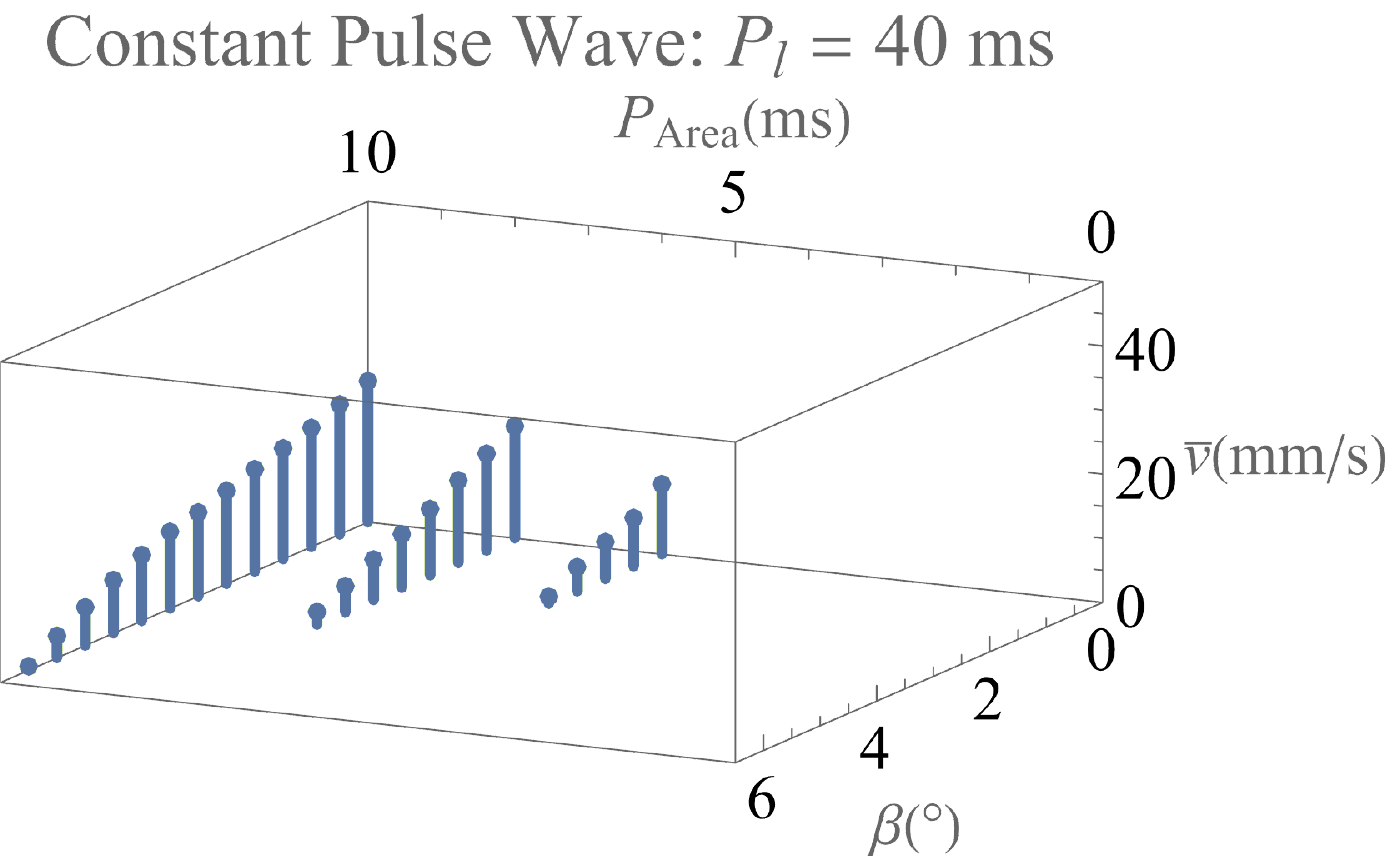}
\end{subfigure}
\hfill
\begin{subfigure}{\widthy}
        \centering
	\includegraphics[width=\widthyt]{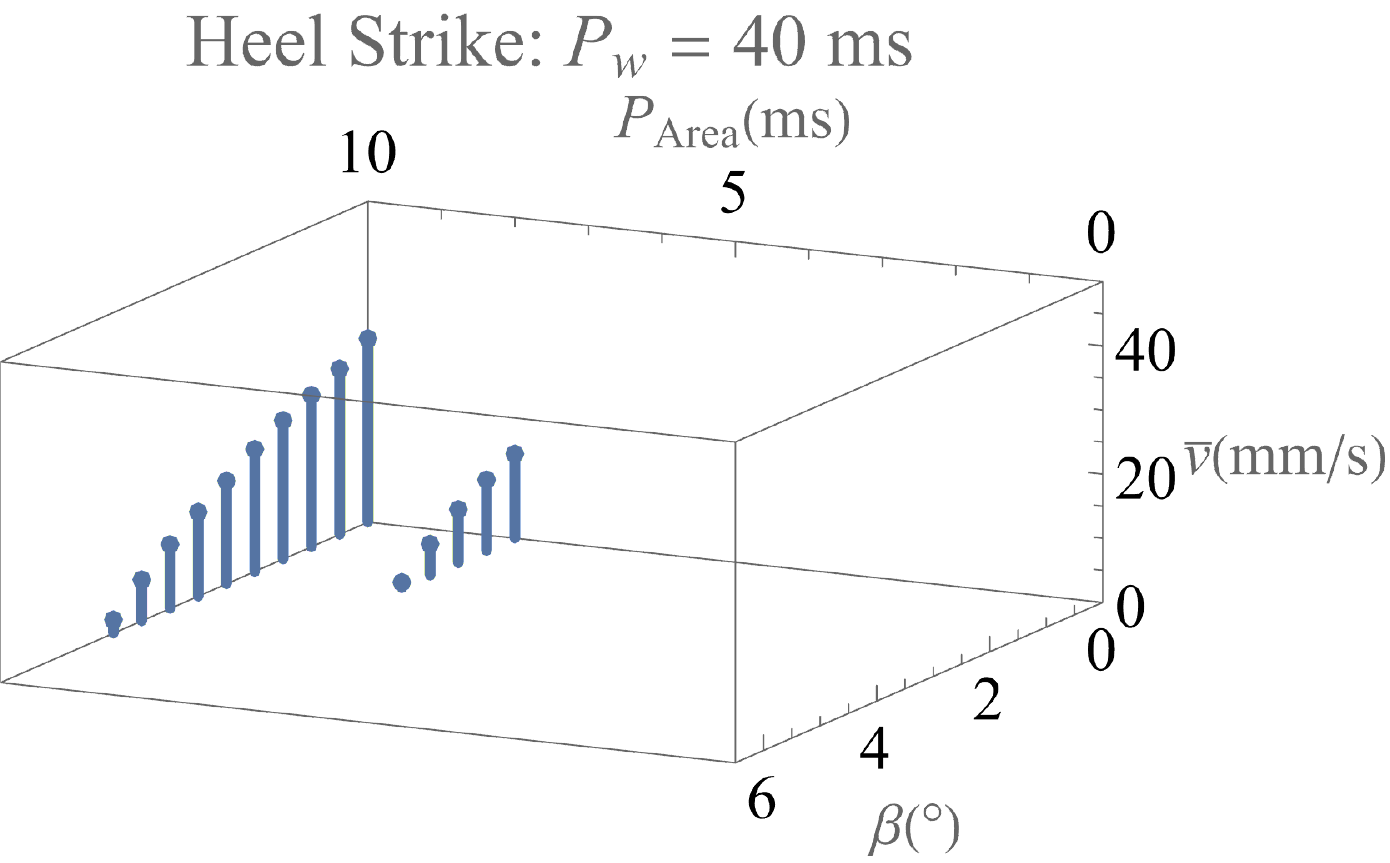}
\end{subfigure}
\caption{Simulation velocities\label{fig:VelocityComposite}}
\end{figure*}

\begin{figure*}[h!]
\begin{subfigure}{\widthy}
        \centering
	\includegraphics[width=\widthyt]{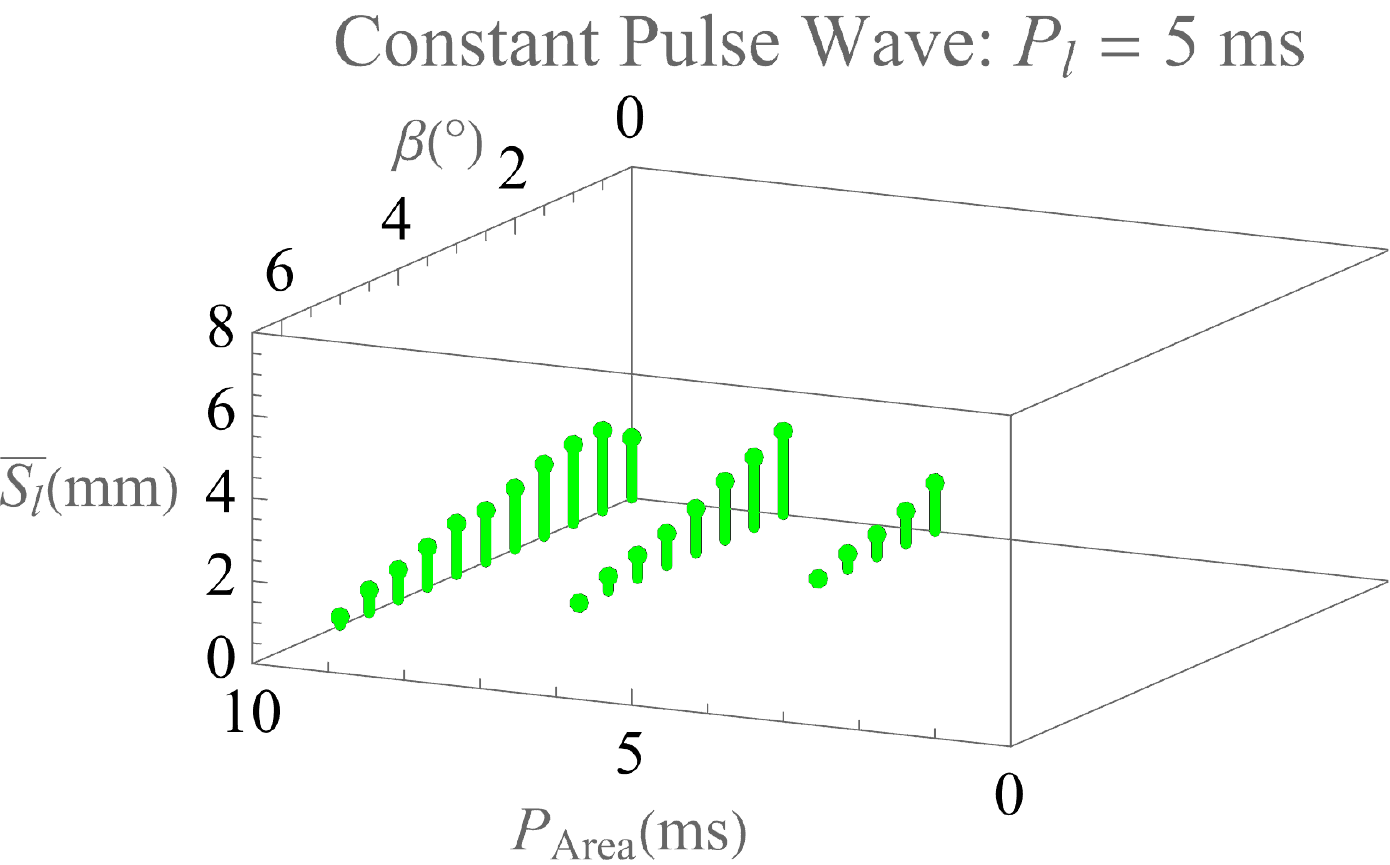}
\end{subfigure}
\begin{subfigure}{\widthy}
        \centering
	\includegraphics[width=\widthyt]{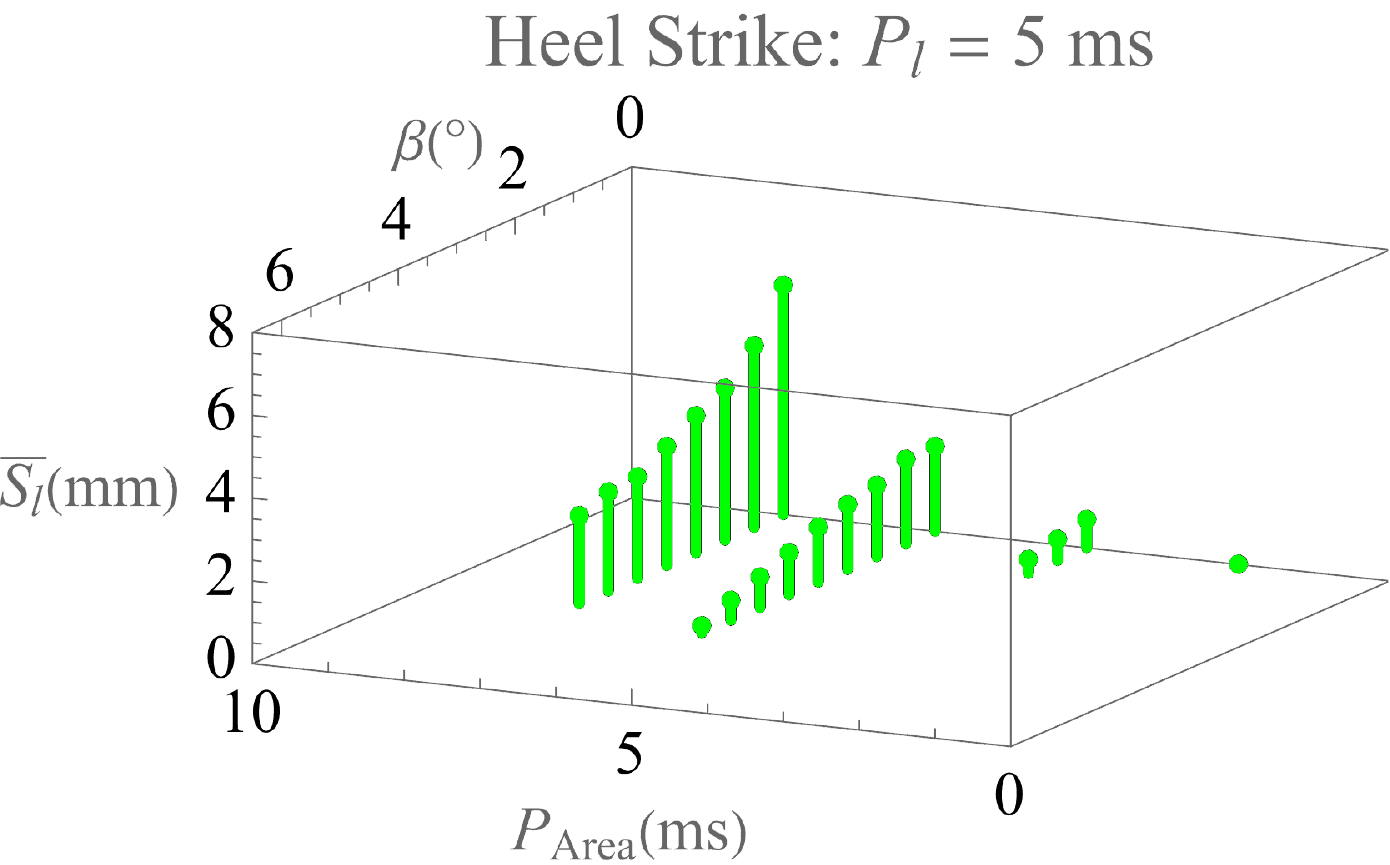}
\end{subfigure}
\begin{subfigure}{\widthy}
        \centering
	\includegraphics[width=\widthyt]{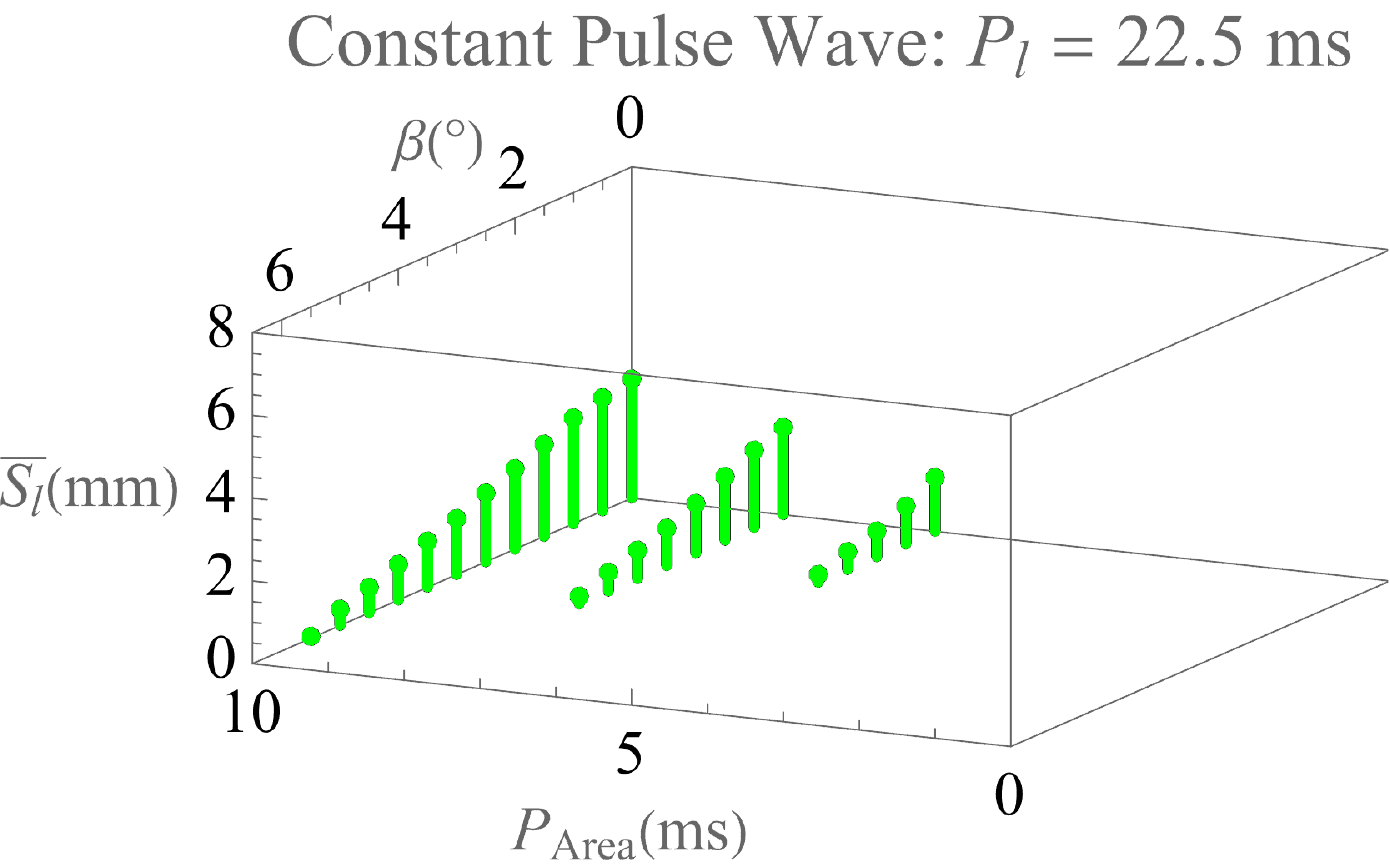}
\end{subfigure}
\begin{subfigure}{\widthy}
        \centering
	\includegraphics[width=\widthyt]{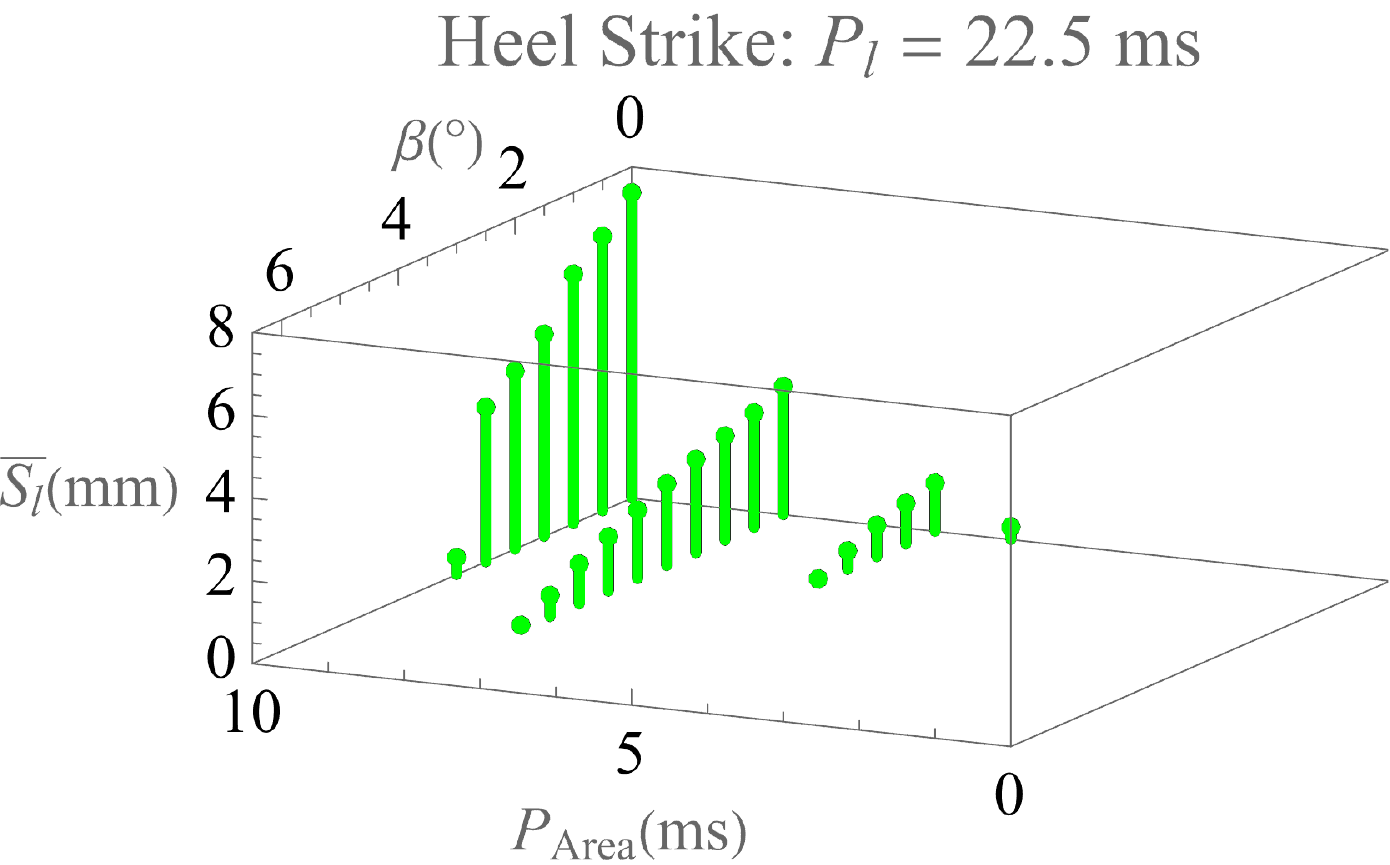}
\end{subfigure}
\begin{subfigure}{\widthy}
        \centering
	\includegraphics[width=\widthyt]{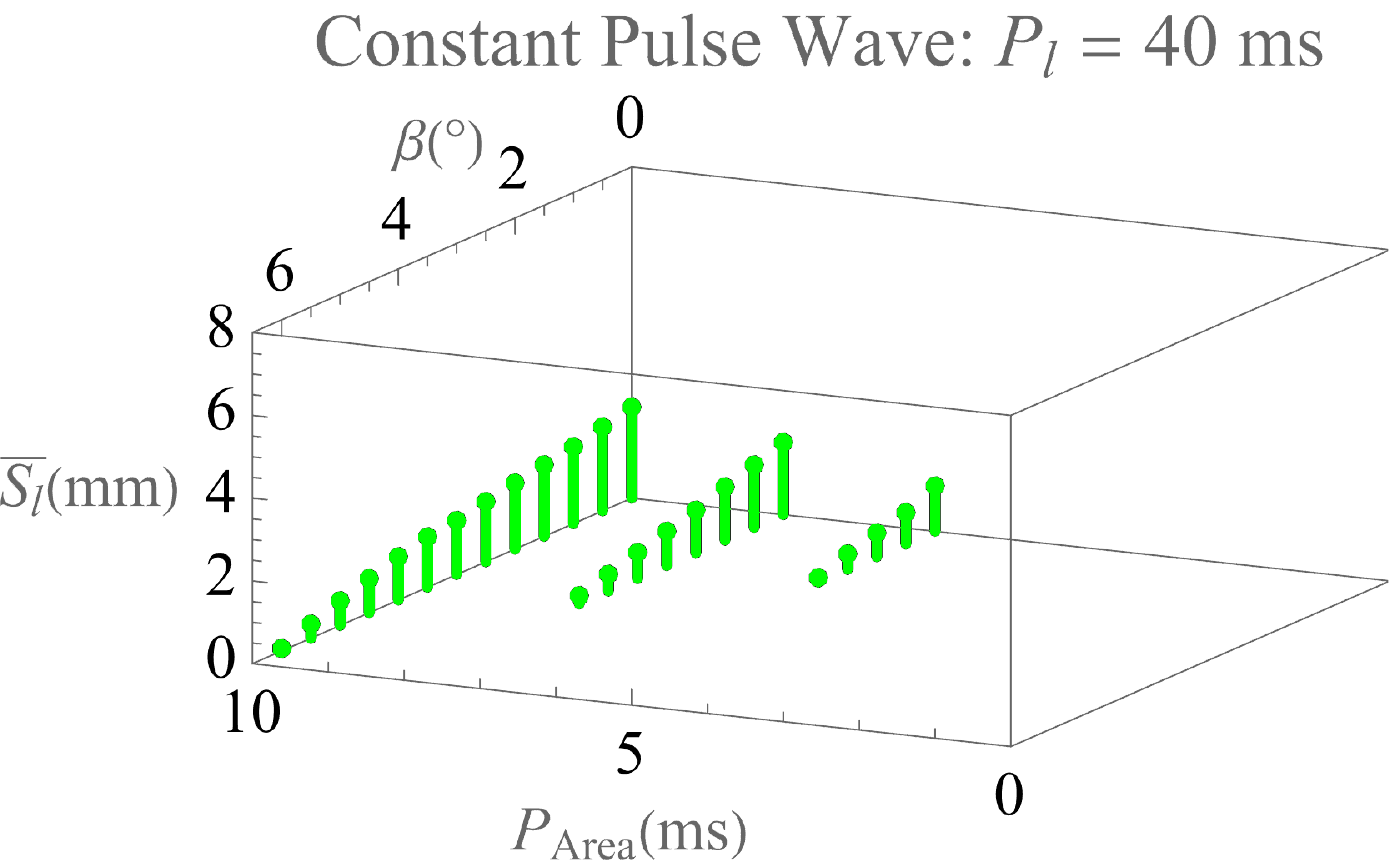}
\end{subfigure}
\hfill
\begin{subfigure}{\widthy}
        \centering
	\includegraphics[width=\widthyt]{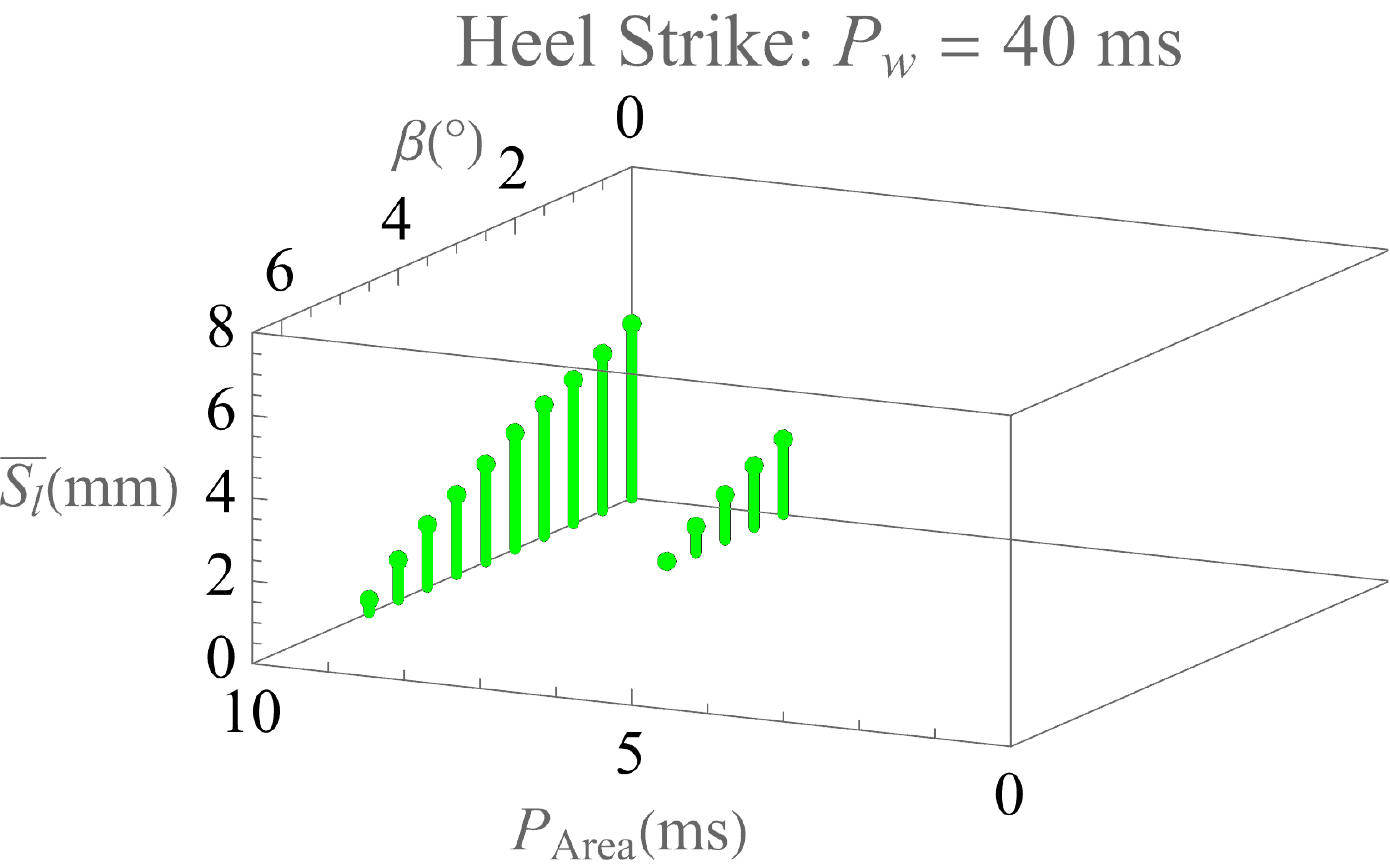}
\end{subfigure}
\caption{Simulation Stride Lengths\label{fig:StrideLengthComposite}}
\end{figure*}

Another performance metric of Big Foot is the locomotion travel, or more specifically the velocity and stride length of the gaits. In Fig.~\ref{fig:VelocityComposite} and Fig.~\ref{fig:StrideLengthComposite}, these metrics are presented. To find the velocity and stride length, each fixed-point is run for 10 seconds. Then, the velocity is calculated by dividing the distance covered by 10 seconds. The stride length is calculated by dividing the distance covered by the number of steps taken. These calculations provide averages, which are required as many of the fixed-points were not period-1.



The results of the study of the travel show the velocities and stride lengths match in their relationship to slope, pulse area, and pulse duration. Also, the velocities and stride lengths appear to be independent of the gait patterns generated. Finally, the highest travel velocity (50 mm/s) and stride length (8mm) are achieved by the heel strike actuation scheme at $P_l=22.5\text{ ms}$ and $P_{Area}=10\text{ ms}$.

\section{Experimental Setup}
\begin{figure}[]
\centering
\includegraphics[width=0.8\columnwidth]{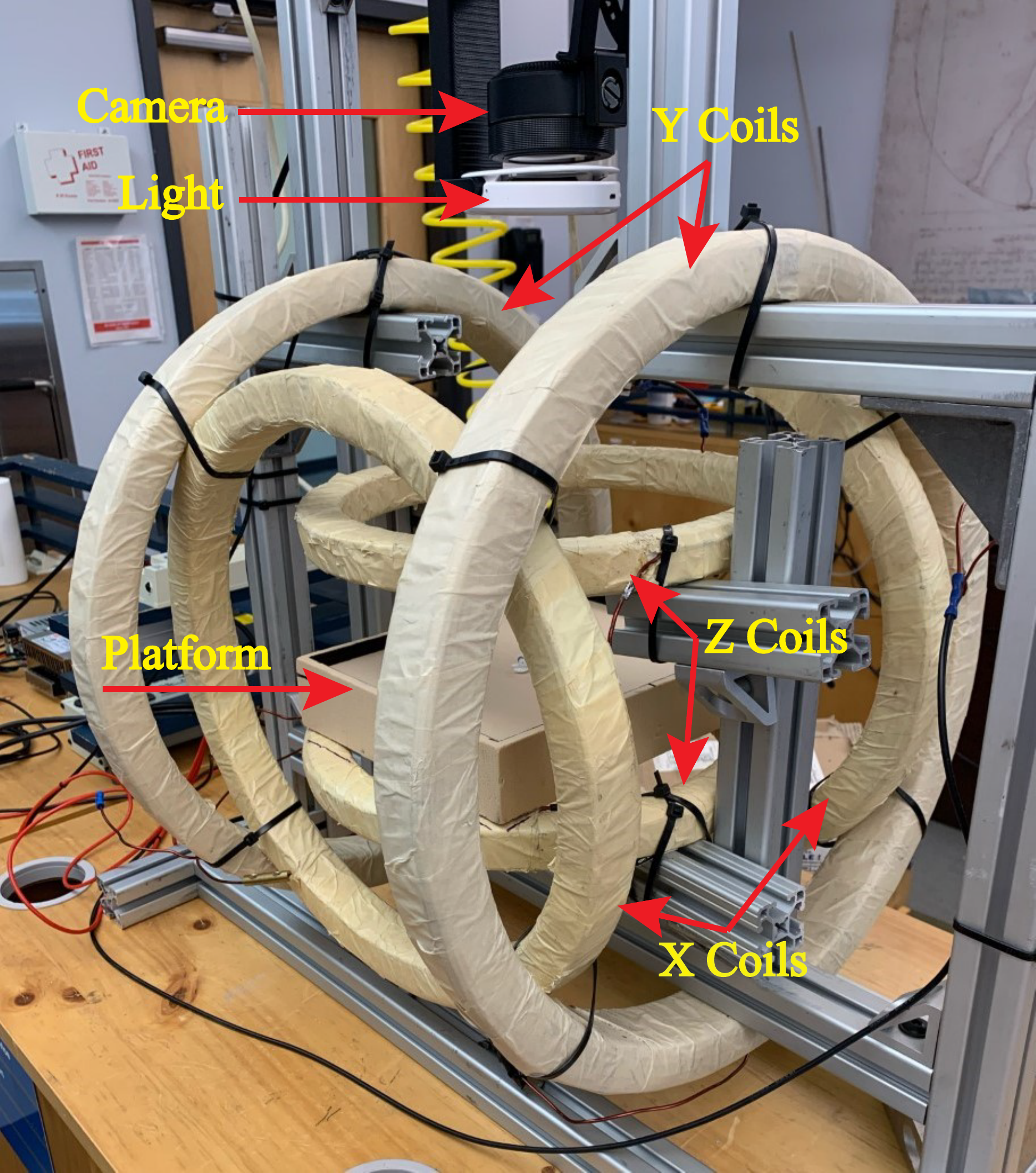}
\caption{Experimental setup of three orthogonally positioned pairs of Helmholtz coils\label{fig:Setup_MBD}}
\end{figure}

The experimental setup includes 3 pairs of Helmholtz coils, a test platform, a light, a camera, a Microcontroller, a motor driver, and a PC. Each pair of Helmholtz coils requires two identical, parallel coils separated by a distance equal to their radius. Both coils are wired such that their current runs in the same direction. The experimental setup follows the outline set by \cite{Abbott2015} for nested circular coils that was used by \cite{al_khatib_razzaghi_hurmuzlu} as well. The setup involves three pairs of Helmholtz coils positioned orthogonally around a platform on which Big Foot walks, as pictured in Fig.~\ref{fig:Setup_MBD}

The outer diameters of the $y$, $x$, and $z$ coils are approximately 46 cm, 40 cm, and 31 cm respectively. The inner diameters are approximately 40 cm, 34 cm, and 25.5 cm and the coil thicknesses are approximately 4.47 cm, 3.61 cm, and 3.21 cm. The platform is machined out of engineering foam, and an Ultra-Thin 10A durometer Silicone Rubber Sheet is placed on top of the platform's surface to prevent slipping. At the top of the setup a Razer Kiyo Pro camera is secured with a XJ-19 Selfie Ring Light around it. Reflective tape is applied to the Shaft Body of Big Foot. Testing of various incline angles is done with the use of 3D printed ramps that hooked on the bottom $z$ coil. 


The approximately uniform magnetic fields caused by the three pairs of coils induce torques, as briefly noted in Eq.~\ref{MagneticMoment}. The input control of the coils contains three variables that are the attitude angles and power of the overall created magnetic field. These variables are: yaw ($\psi_m$), pitch ($\phi_m$), and power($P$). The equations used to convert these inputs into actual coil voltages are given as follows:

\begin{equation} \label{MagneticAnglesToPowers}
\begin{split}
P_x & = P \cos{\phi_m}\cos{\psi_m}k_x \\
P_y & = P \cos{\phi_m}\sin{\psi_m}k_y \\
P_z & = P \sin{\phi_m}k_z
\end{split}
\end{equation}
where, $\{P_x,P_y,P_z\}$ are the separate voltages applied to each orthogonal coil and $\{k_x,k_y,k_z\}$ are the correction factors that account for unit conversions and intermediate experimental parameters.



\subsection{Big Foot Prototype}

\begin{figure}[]
\centering
\includegraphics[width=6 cm]{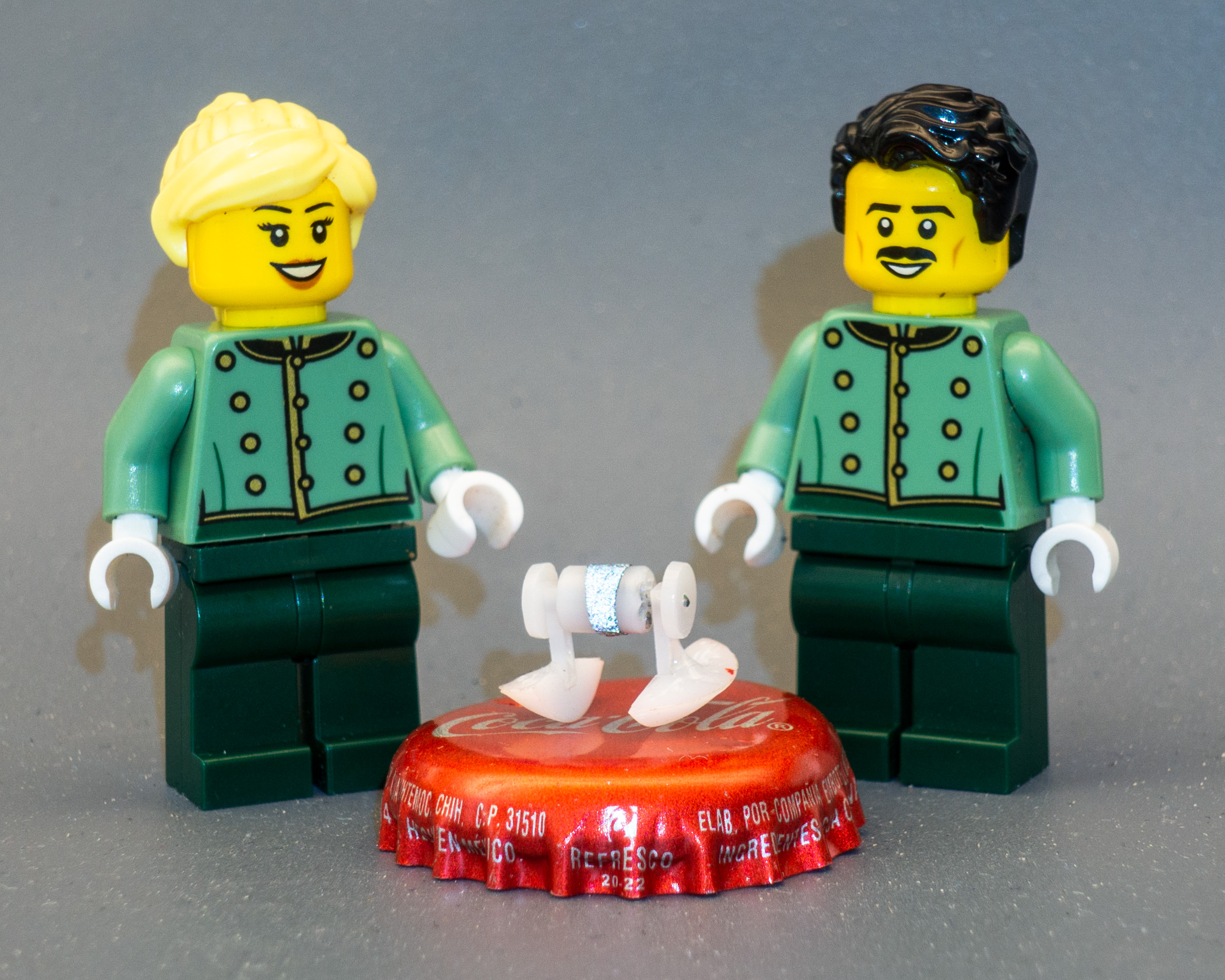}
\caption{The Big Foot\label{BigFootsPortrait}}
\end{figure}  

The prototype of Big Foot is shown in Fig.~\ref{BigFootsPortrait}. The Shaft Cylinder is machined from Delrin. The collars and legs were laser cut using the PS48 laser cutter and engraver. The leg is 6.5x0.79x1.75 mm. The feet are 3D printed using a Photon S SLA 3D printer. The feet are a section of an 8mm sphere, which are glued to the legs. The shaft is a 0.794mm non-magnetic stainless steel shaft. The magnets used are 1mm diameter Neodymium magnets.

\section{Experimental Results}
A set of walking tasks is defined to validate the performance of the Big Foot prototype. First, Big Foot is placed along a straight path with different slopes to compare the prototype's dynamics to the model. Second, a variety of two dimensional paths are prescribed with Big Foot to demonstrate maneuvering capabilities.

\subsection{Uphill Walking Experiments}
To compare the response of the Big Foot prototype to the model, the prototype is placed on a series of slopes. As no sensors were available to detect the heel strikes, experiments are limited to the constant pulse wave actuation scheme. The parameters of this test were as follows:$\{\psi_m=20^\circ,\phi_m=60^\circ,P_m=33\%,P_L=300\text{ ms},T_{off}=60\text{ ms}\}$. A long pulse width is used because a very short pulse lengths are not possible because the cut off frequency of the largest Helmholtz coil is found experimentally to be 5Hz. Video footage of the experiment is shown in \href{https://www.youtube.com/@smusystemslaboratory/playlists}{Extension 1}.

\begin{figure}[]
\centering
\includegraphics[width=\columnwidth]{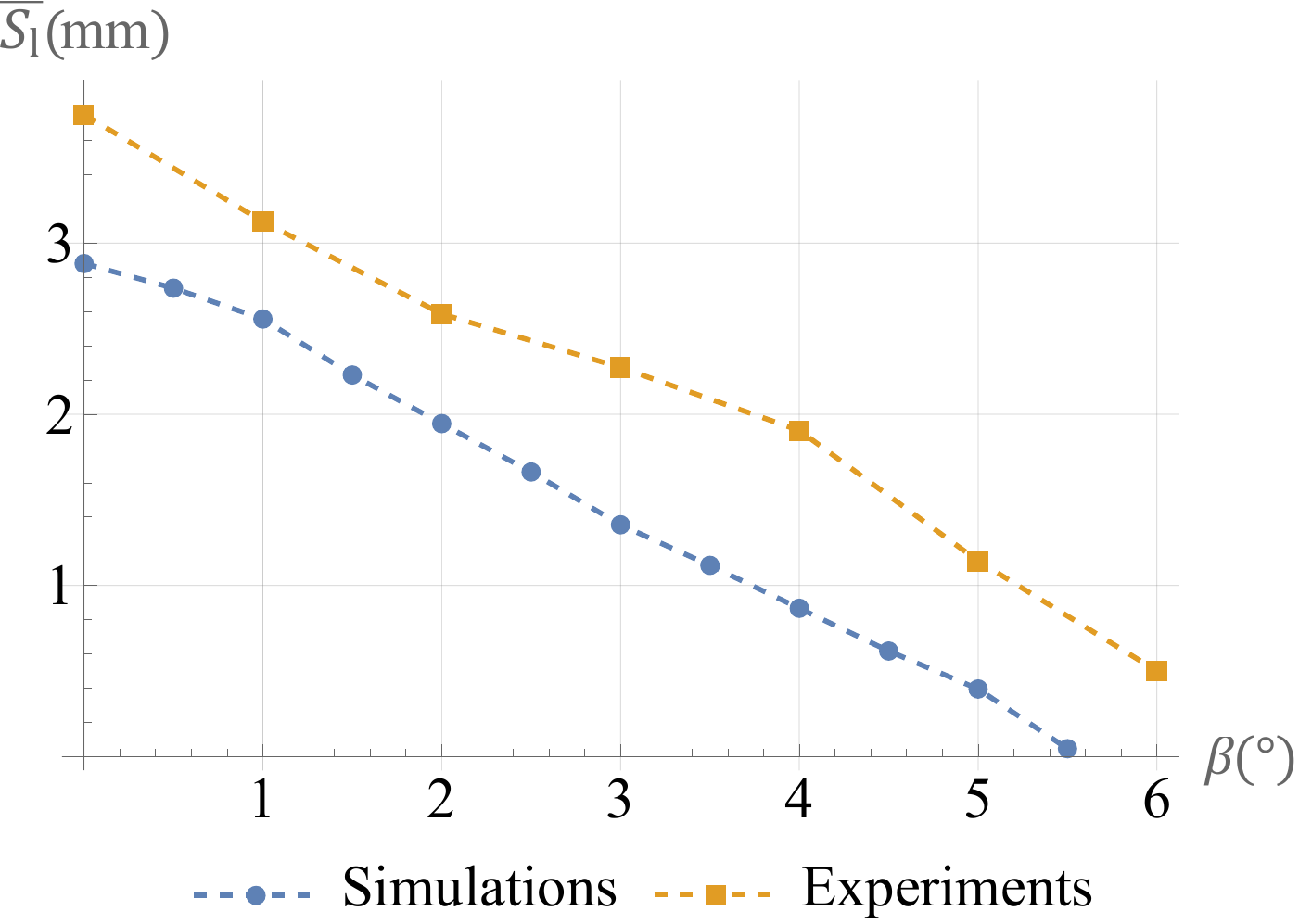}
\caption{Experimental results compared to Simulations.\label{ExperimentalVsSimulation}}
\end{figure} 

Figure~\ref{ExperimentalVsSimulation} shows the stride length's variation with respect to the walking surface slope for both experiments and simulations. The simulation results agree with the experiments. Experimental outcomes are similar to the theoretical ones. The stride length is greater in the experiment, but this can be explained in experimental uncertainties such as manufacturing tolerances and inconsistent surface contact. However, the result in Fig.~\ref{ExperimentalVsSimulation} are consistent with other runs and show the same relationship in experiments and simulations. In constant pulse wave actuation, the stride length, and consequentially the forward velocity, monotonically decreases as the walking surface slope increases.

\begin{figure}[]
\centering
\includegraphics[width=\columnwidth]{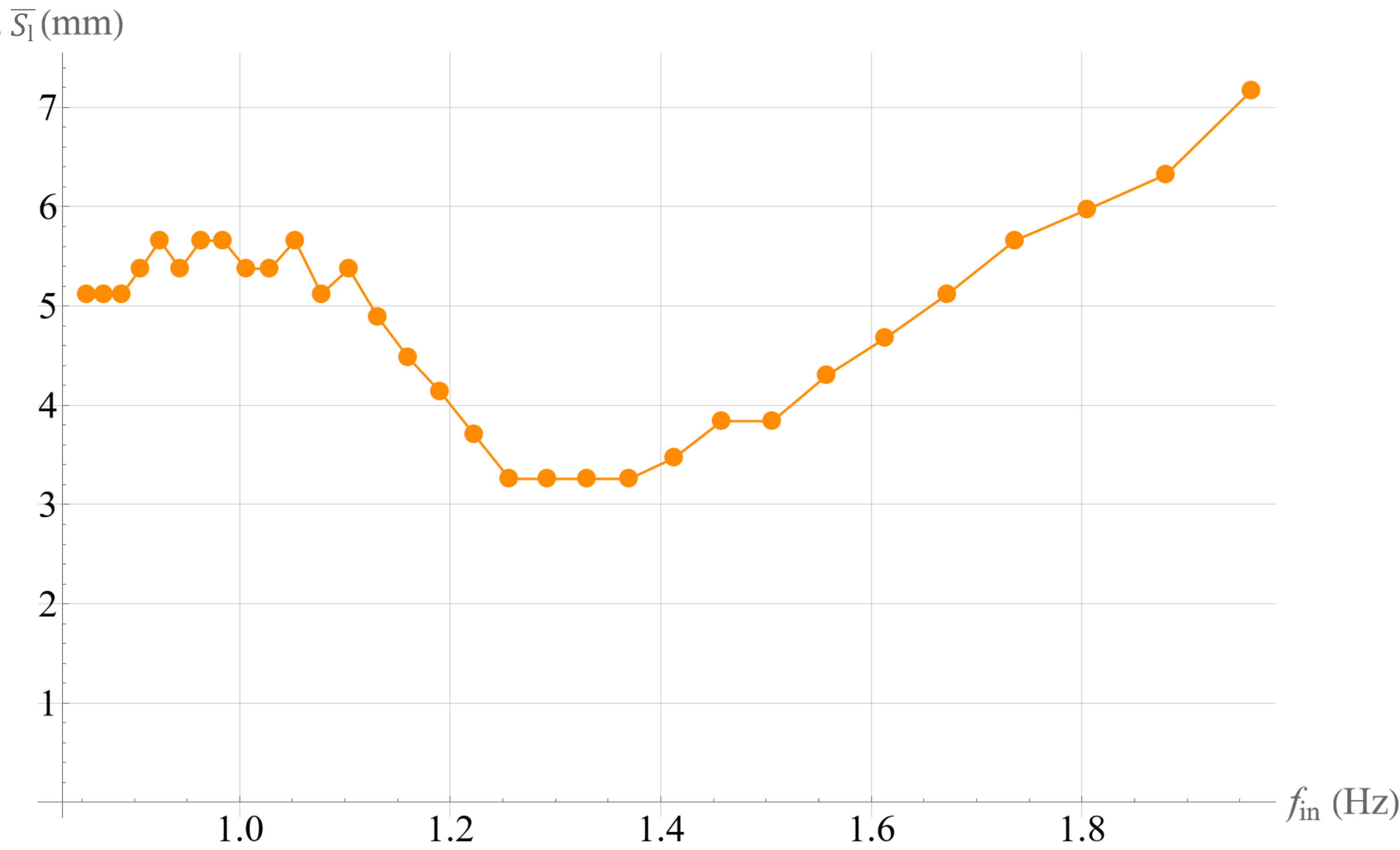}
\caption{Stride Length with respect to input frequency\label{StrideLengthVsFrequency}}
\end{figure}

The next experiment is designed to capture the effect of pulse frequency on the stride length. In the experiment with results in Fig.~\ref{StrideLengthVsFrequency}, the prototype walks across a flat surface. The input pulse duration is incremented by 10\% between each run.

We observe two different gait characteristics at two extremes of the input frequencies. For high frequencies the gait is more dynamic because of the short duration pulses. For low frequencies, however, the gait becomes more quasi-static. This is due to the long duration pulse where the biped becomes almost static towards the end of each step cycle. When the contra-lateral pulse arrives, the biped transitions to the next quasi-static posture, achieving forward locomotion. For this reason, for lower frequency values, step length remains almost unchanged. There is a clear transition from impulsive actuation to continuous actuation as the input frequency exceeds a certain value (approximately 1.4 Hz). Experimentally we observe that the longest step lengths are achieved at the highest input frequencies. Consequently, we observe that impulsive actuation produces longer step lengths than continuous actuation.

\subsection{Miscellaneous Maneuvers}
In this paper, so far we have focus on locomotion along a single direction. It is possible to walk in two dimensions by adding a new input parameter through the following substitution:
\begin{equation}
    \psi_m\rightarrow\psi_m+\psi_d
\end{equation}
where, $\psi_d$ is a direction angle.

\begin{figure}[]
\begin{subfigure}{\widthy}
        \centering
	\includegraphics[width=0.935\linewidth]{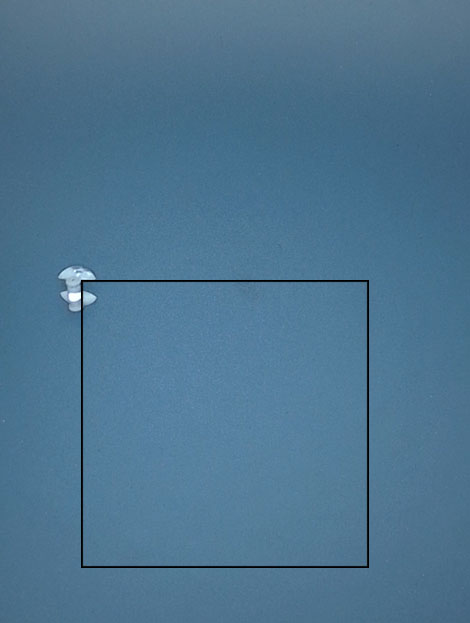}
 \caption{Square Path}
\end{subfigure}
\begin{subfigure}{\widthy}
        \centering
	\includegraphics[width=\linewidth]{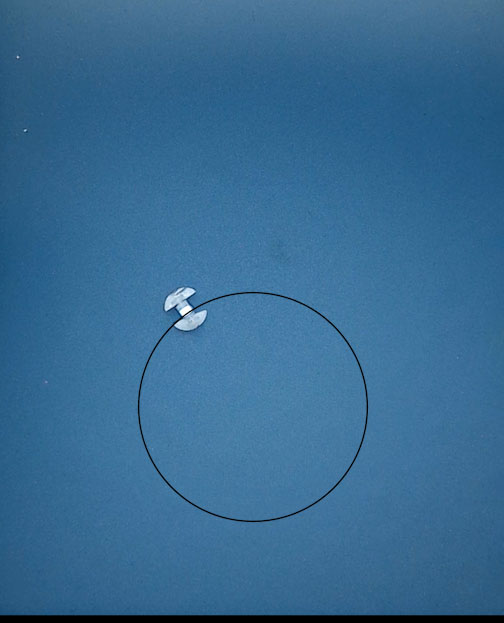}
 \caption{Circular Path}
\end{subfigure}
\begin{subfigure}{\widthy}
        \centering
	\includegraphics[width=\linewidth]{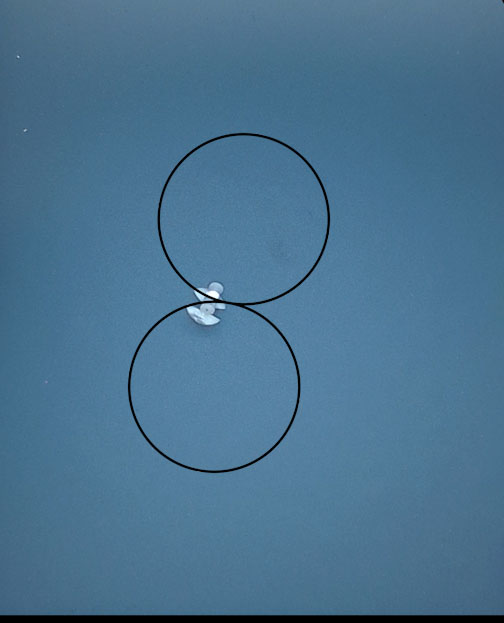}
 \caption{Figure-8 Path}
\end{subfigure}
\hfill
\begin{subfigure}{\widthy}
        \centering
	\includegraphics[width=\linewidth]{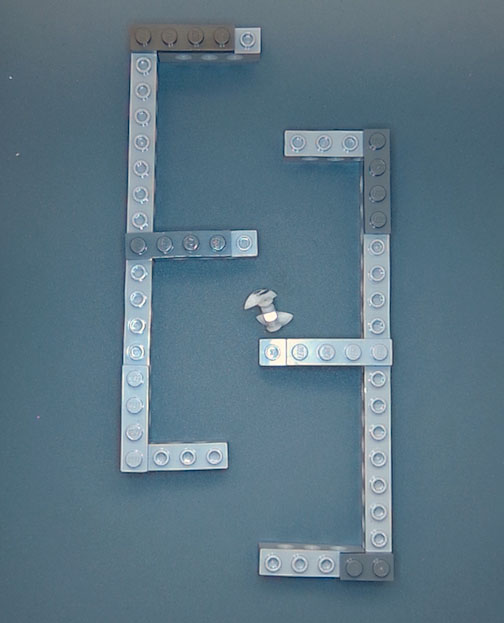}
 \caption{Lego\textsuperscript{\textregistered} Maze}
\end{subfigure}
\caption{2D Locomotion Paths\label{fig:2DPaths}}
\end{figure}

Big Foot can be steered using $\psi_d$. Each time a pulse is administered a magnetic field is generated in a given direction. This results in an applied moment vector in that direction. Hence, the locomotion direction can be altered by varying this $\psi_d$ angle. In order to demonstrate this feature, four sets of maneuvers were performed. These sets can be viewed in \href{https://www.youtube.com/@smusystemslaboratory/playlists}{Extension 2-5}. Figure~\ref{fig:2DPaths} depicts the four tasks that we put into Big Foot.

The first maneuver is programmed to have Big Foot follow a square path. The maneuver is simply 36 forward steps, followed by a $90^\circ$ left turn. The turn is accomplished by adding $90^\circ$ to $\psi_d$. The sequence is repeated until the square is complete. This initial maneuver shows the ability of Big Foot to make sharp right turns.

The second maneuver is a circular path. Here, Big Foot takes 80 steps. This maneuver is realized by taking alternating steps with $\psi_d = \psi_d + 4^\circ$ in one step and $\psi_d = \psi_d + 5^\circ$ in the next one.

The third maneuver is a figure eight. It is realized similarly to the circle. Except, after one loop, one accomplishes the second loop by subtracting from $\psi_d$ instead of adding to it.

The final maneuver is a maze. Lego\textsuperscript{\textregistered} bricks are used in assembling the maze. This maneuver is accomplished using right turns and straight walks similar to the square maneuver.

These maneuvers show the practical potential of hip actuated bipedal locomotors. The walking direction can be easily modified by inducing hip rotations. The work in this paper has shown that gait regulation can be achieved through hip action.

\section{Conclusion}
The work in this paper presents a new magnetically actuated bipedal walker: Big Foot. Two important questions regarding bipedal locomotion are answered. First, Big Foot has shown that bipedal locomotion and maneuvering can be achieved through pure hip actuation. Second, the analysis shows that it is preferable to used impulsive actuation rather than continuous actuation because it results in longer step lengths. In addition, the performance of this walker is also analyzed through simulation. After defining two distinct actuation schemes; heel strike and continuous pulse wave actuation schemes. It is found that heel strike actuation achieves better stability, more predictable gait generation, and greater locomotion velocities. However, constant pulse wave actuation achieves locomotion on steeper slopes. What is also observed is that, even though increasing the pulse duration in heel strike increases the maximum slope, it also requires more magnetic power. Interestingly more magnetic power in heel strike results in shorter stride lengths as the pulse duration is increased.  The work opens the door for experimentation with new designs of bipedal walkers using hip actuation (magnetic and traditional actuation). The work also shows that magnetic actuation promises to speed up locomotion research as prototypes are no longer limited by financial resources and production time. Future research includes studying more complex legged machines and gait patterns. We will explore tasks that include stair climbing, obstacle avoidance, and closed-loop control.


%





\begin{acks}
We would like to thank Kenny Sangston and Necdet Yildirimer for providing machining services and space in manufacturing Big Foot. We would also like to thank The SMU Deason Innovation Gymnasium, led by JT Ringer and Seth Orsborn, for allowing us to use their laser cutter. Finally, we are acknowledging Ying-Chu Chen for her work in the development of prototypes that led to the idea of hip actuation.
\end{acks}


\bibliographystyle{SageH}
\bibliography{Cox_Beskok_Hurmuzlu_Big_Foot.bib}

\section{Appendix A: Index to multimedia Extensions}
Table of Multimedia Extensions.
\begin{table}[H]
\begin{tabular}{lll}
\hline
Extension & Media Type & Description   \\ \hline
1         & Video      & Walking Uphill Increasing Slopes       \\
2         & Video      & Walking a Square Path   \\
3         & Video      & Walking a Circular Path \\
4         & Video      & Walking a Figure-8      \\
5         & Video      & Navigating a Lego\textsuperscript{\textregistered} Maze     \\ \hline
\end{tabular}
\end{table}





\end{document}